\documentclass[10pt]{article}

\usepackage[preprint]{bakelab}

\usepackage{xspace}
\usepackage{colortbl}
\usepackage{pifont}
\usepackage{placeins}
\usepackage{multirow}
\usepackage{wrapfig}
\usepackage{nicefrac}
\usepackage{fontawesome5}

\definecolor{narragreen}{HTML}{2C3B1F}      % olive black-green
\definecolor{narragreendeep}{HTML}{1A2412}  % deeper, for bold heads

\definecolor{ccuda}{HTML}{DEC89A}           % CUDA — dusty gold
\definecolor{ccudasoft}{HTML}{ECDDA6}       % CUDA outer
\definecolor{cmodel}{HTML}{E3B3C5}          % Model Development — rose
\definecolor{cmodelsoft}{HTML}{F2DAE0}      % Model Development outer
\definecolor{cpuzzle}{HTML}{D9AE9A}         % Puzzle \& Challenge — terracotta
\definecolor{cpuzzlesoft}{HTML}{ECD6C8}     % Puzzle \& Challenge outer
\definecolor{csystem}{HTML}{B3D196}         % System Optimization — sage
\definecolor{csystemsoft}{HTML}{C9DCB1}     % System Optimization outer

\definecolor{sagegreen}{HTML}{6B8F71}
\definecolor{citegreen}{HTML}{8E9C5E}  % citation green — olive/sage
\definecolor{olivegreen}{HTML}{708238}

\hypersetup{
  colorlinks=true,
  citecolor=citegreen,
  linkcolor=sagegreen,
  urlcolor=olivegreen,
}

\newcommand{\bench}{\textsc{AutoLab}\xspace}
\newcommand{\phead}[1]{\noindent\textcolor{narragreendeep}{\textbf{#1}}\ }

\newcommand{\corres}{\textsuperscript{\kern-0.05em\color{BakeAccent}\scalebox{0.72}{\faEnvelope[regular]}}}

\newcommand{\authgap}{\hspace{1.5em}}

\newcommand{\reslink}[4]{\href{#1}{\textcolor{BakeDark}{#2}\,\textcolor{BakeAccent}{\texttt{#3}}\ \textcolor{BakeGray}{\texttt{#4}}}}
\newcommand{\ressep}{\hspace{0.8em}{\color{BakeSubtle}\rule[-0.25ex]{0.5pt}{1.5ex}}\hspace{0.8em}}

\newtcolorbox{taskcard}[3]{%
  enhanced, breakable, sharp corners,
  colback=#1!18!white, colframe=#1!85!black,
  boxrule=0pt, leftrule=2.5pt,
  left=8pt, right=6pt, top=3pt, bottom=3pt,
  before skip=3pt, after skip=3pt,
  before upper={\textbf{\texttt{#2}}\hfill\textit{\small\textcolor{black!60}{#3}}\par\smallskip}
}

\newtcolorbox{promptbox}[1]{%
  enhanced, breakable, sharp corners,
  colback=csystem!12!white, colframe=csystem!75!black,
  boxrule=0.4pt, leftrule=2.4pt,
  left=8pt, right=8pt, top=6pt, bottom=6pt,
  before skip=6pt, after skip=6pt,
  fonttitle=\bfseries\small,
  coltitle=narragreendeep, colbacktitle=csystem!28!white,
  attach boxed title to top left={xshift=10pt, yshift*=-\tcboxedtitleheight/2},
  boxed title style={sharp corners, boxrule=0pt, colframe=csystem!75!black,
    left=4pt, right=4pt, top=1.5pt, bottom=1.5pt},
  title={#1},
}

\newcommand{\del}[1]{}

\title{\bench{}: Can Frontier Models Solve Long-Horizon Auto Research and Engineering Tasks?}

\author{%
\mbox{Zhangchen Xu$^{1,11}$\equalcontrib}\authgap
\mbox{Junda Chen$^{4}$}\authgap
\mbox{Yue Huang$^{5}$}\authgap
\mbox{Dongfu Jiang$^{8,10}$}\authgap
\mbox{Jiefeng Chen$^{12}$}\\[3pt]
\mbox{Hang Hua$^{13}$}\authgap
\mbox{Zijian Wu$^{7}$}\authgap
\mbox{Zheyuan Liu$^{5}$}\authgap
\mbox{Zexue He$^{2}$}\authgap
\mbox{Lichi Li$^{14}$}\\[3pt]
\mbox{Shizhe Diao$^{10}$}\authgap
\mbox{Jiaxin Pei$^{2}$}\authgap
\mbox{Jinsung Yoon$^{12}$}\authgap
\mbox{Hao Zhang$^{4}$}\authgap
\mbox{Mengdi Wang$^{6}$}\\[3pt]
\mbox{Radha Poovendran$^{1}$}\authgap
\mbox{Misha Sra$^{3}$}\authgap
\mbox{Alex Pentland$^{2,9}$}\authgap
\mbox{Zichen Chen$^{2,3,11}$\equalcontrib\corres}%
}

\affiliations{%
$^{1}$University of Washington \quad
$^{2}$Stanford University \quad
$^{3}$UCSB \quad
$^{4}$UCSD \quad
$^{5}$University of Notre Dame \\[1pt]
$^{6}$Princeton University \quad
$^{7}$NUS \quad
$^{8}$University of Waterloo \quad
$^{9}$MIT \quad
$^{10}$NVIDIA \quad
$^{11}$Bake AI \quad
$^{12}$Google \\[1pt]
$^{13}$MIT-IBM Computing Research Lab \quad
$^{14}$Independent Researcher \\[3pt]
{\footnotesize\itshape \textcolor{BakeAccent}{$^{*}$}Equal contribution. \quad \textcolor{BakeAccent}{\scalebox{0.85}{\faEnvelope[regular]}}\,Corresponding author.}%
}

\paperdate{\today}

\begin{document}

\maketitle

% ============================================================================
%  Content
% ============================================================================
\begin{abstract}
Scientific and engineering progress is fundamentally a long-horizon iterative process: proposing changes, running experiments, measuring outcomes, and continuously refining artifacts. Yet existing benchmarks for frontier models primarily evaluate either single-turn responses or short-horizon agent trajectories, failing to capture the challenges of sustained iterative improvement over extended time horizons.
To address this gap, we introduce \bench{}, a new benchmark for ultra long-horizon closed-loop optimization. \bench{} consists of 36 realistic, expert-curated tasks spanning four diverse domains: system optimization, puzzle \& challenge, model development, and CUDA kernel optimization. Each task begins with a correct but deliberately suboptimal baseline and challenges agents to improve it within a strict wall-clock budget.
Evaluating 17 state-of-the-art models reveals the dominant predictor of success is not the quality of an agent’s initial attempt, but its persistence in repeatedly benchmarking, editing, and incorporating empirical feedback. While \texttt{claude-opus-4.6} exhibits strong long-horizon optimization capabilities, most frontier models, including several proprietary ones, either terminate prematurely or exhaust their budgets with minimal progress. These results underscore the importance of time awareness and persistent iteration in autonomous agents.
We open-source the full benchmark, evaluation harness, and task artifacts, to accelerate research toward truly capable long-horizon agents.

\vspace{6pt}
{\raggedright\reslink{https://github.com/autolabhq/autolab}{\faGithub}{Code}{autolabhq/autolab}\ressep\reslink{https://autolab.moe/}{\faGlobe}{Website}{autolab.moe}\par}
\end{abstract}

\begin{figure}[ht!]
  \centering
  \includegraphics[width=0.95\linewidth]{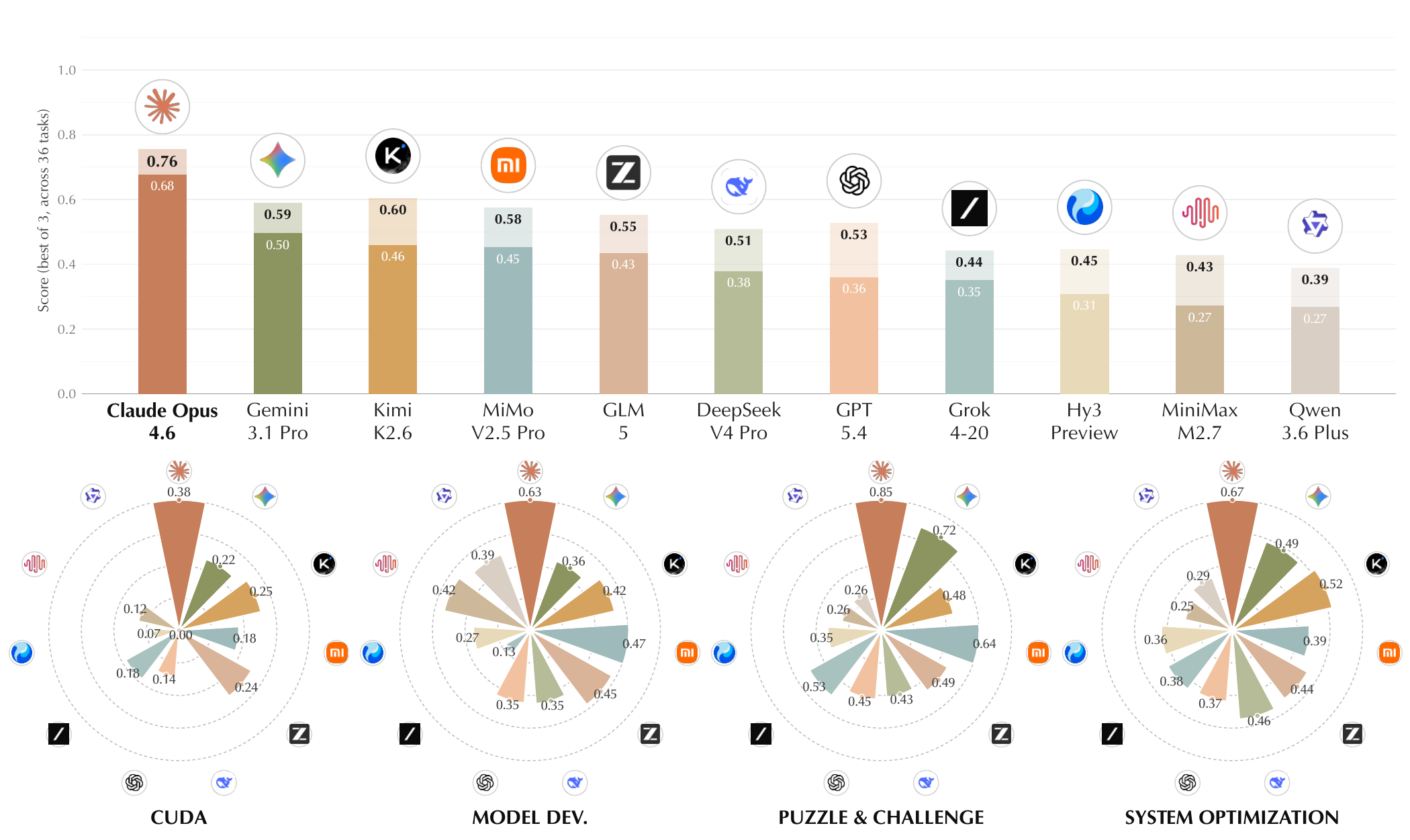}
  \caption{\bench{} benchmarks frontier models across 36 tasks spanning 4 categories, here for the 11 provider flagships (one model per provider). \emph{Top:} models sorted left-to-right by Avg@3 (solid bar); the translucent extension reaches Best@3. \emph{Bottom:} four rose charts, one per category, where each petal's length is that model's Avg@3 in the category. \texttt{claude-opus-4.6} leads all four categories and the overall ranking; the runner-up rotates by category.}
  \label{fig:teaser}
\end{figure}

\section{Introduction}

Frontier LLM agents are increasingly deployed on tasks that play out over hours rather than minutes, from post-training models~\citep{rank2026posttrainbench} and optimizing low-level systems~\citep{chi2026frontier} to running open-ended research loops~\citep{novikov2025alphaevolve, karpathy2026autoresearch}. Progress on such tasks is iterative: it comes from inspecting an artifact, proposing a change, running experiments, measuring the outcome, and refining over many cycles, not from a single correct answer. Sustaining this loop over a long horizon requires managing time, compute, and noisy empirical signals. Short, single-shot evaluations are not designed to test whether today's frontier models can do so.

Current evaluations largely overlook this regime. Static, single-turn coding benchmarks primarily test model knowledge and one-shot coding~\citep{jain2025livecodebench, zhuo2025bigcodebench}. Another wave of agentic benchmarks has extended to short, interactive trajectories~\citep{mialon2023gaia, liu2023agentbench, jimenez2024swebench, merrill2026terminalbench}. Only lately have a few benchmarks begun to explore hour-long, closed-loop optimization~\citep{ouyang2025kernelbench, nathani2025mlgym, mang2025frontiercs, lupidi2026airsbench}. However, these efforts remain limited in both scale and generality.

Two major obstacles have kept sustained long-horizon optimization progress slow. First, the most impressive demonstrations of empirical optimization, such as AlphaEvolve~\citep{novikov2025alphaevolve} and ~\citeauthor{karpathy2026autoresearch}’s AutoResearch agent~\citep{karpathy2026autoresearch}, are tightly coupled to heavily engineered, model-specific harnesses, tools, and search strategies. This co-design makes it difficult to isolate the underlying model’s true contribution. Second, existing long-horizon benchmarks are narrow in scope, each targeting a single domain such as ML engineering~\citep{rank2026posttrainbench, starace2025paperbench}, systems and kernel optimization~\citep{ouyang2025kernelbench, mang2025frontiercs}, or real-world engineering~\citep{chi2026frontier}. Crucially, none of these benchmarks simultaneously offer broad coverage across scientific and engineering domains while maintaining high difficulty and resistance to saturation.

To close this gap, we introduce \bench{}, a benchmark for ultra long-horizon closed-loop optimization with LLM agents. Each task in \bench{} provides a correct but deliberately suboptimal baseline and challenges the agent to improve it iteratively within a strict wall-clock budget. \bench{} comprises 36 executable tasks across \textbf{four categories}: system optimization, puzzle \& challenge, model development, and CUDA kernel optimization. Its design rests on three commitments: (1) tasks must demand sustained empirical iteration over long horizons; (2) scoring must be continuous and well-calibrated, rewarding partial progress across heterogeneous metrics; and (3) evaluation must be hack-resistant, enforced through sealed evaluators, correctness gates, immutable-file checks, and adversarial auditing.

Long-horizon optimization represents a distinct capability that cannot be reduced to agentic coding ability. This is clearly demonstrated by our main evaluation (Figure~\ref{fig:teaser}), which consumed a total of \textbf{2{,}544} wall-clock hours and \textbf{8.60} billion tokens. \texttt{claude-opus-4.6} leads every sub-domain, reaching an Avg@3 of $0.68$ versus $0.50$ for the next-best model. Many otherwise strong models, including \texttt{gpt-5.4}, fail for reasons unrelated to raw coding ability: some terminate after minimal exploration, while others exhaust their entire budget without producing a valid final solution (Section~\ref{sec:analysis}). Our trajectory analysis further shows that final performance correlates more strongly with \textbf{persistence} than with \textbf{one-shot solution quality}: agents that repeatedly benchmark, edit, and incorporate empirical feedback throughout the trajectory achieve substantially better outcomes. These findings suggest that persistence, time awareness, and empirical search will be central to future autonomous research agents.

In summary, we make the following key contributions:
\begin{itemize}[topsep=0pt, itemsep=2pt, leftmargin=*]
\item \textbf{A high-quality benchmark.} We introduce \bench{}, the first benchmark designed specifically for ultra long-horizon closed-loop optimization across diverse domains.
\item \textbf{Large-scale evaluation.} We conduct a systematic evaluation of 17 state-of-the-art models, including four proprietary frontier models, using a fixed, standardized harness under identical experimental conditions.
\item \textbf{In-depth trajectory analysis and insights.} Through comprehensive analysis of all trajectories (including manual inspection of 302 zero-score rollouts), we reveal key behavioral limitations, most notably a lack of time awareness (premature termination versus budget exhaustion). We further show that the dominant predictor of final performance is not the quality of an agent’s initial solution, but its persistence in iterative refinement.
\end{itemize}

\section{The \bench Benchmark}
\label{sec:design}

We present \bench, a benchmark for evaluating frontier models on research and engineering tasks whose horizons are measured in hours rather than minutes. Its design is organized around three commitments. Tasks must be \emph{ultra long-horizon}, demanding sustained empirical iteration across many cycles rather than a single-shot patch; scoring must be \emph{continuous and calibrated}, going beyond pass/fail to support fine-grained relative comparison across heterogeneous metrics such as runtime, perplexity, and parameter count, and to resist saturation as frontier capabilities advance; and verification must be \emph{hack-resistant}, since performance benchmarks expose a far larger attack surface for shortcuts than patch-style benchmarks. The remainder of this section formalizes the task specification (Sec. \ref{subsec:task_formulation}), describes how tasks are sourced and quality-controlled (Sec. \ref{subsec:construction}), and reports the final composition of \bench (Sec. \ref{subsec:composition}). 

\begin{figure}
    \centering
    \includegraphics[width=1\linewidth]{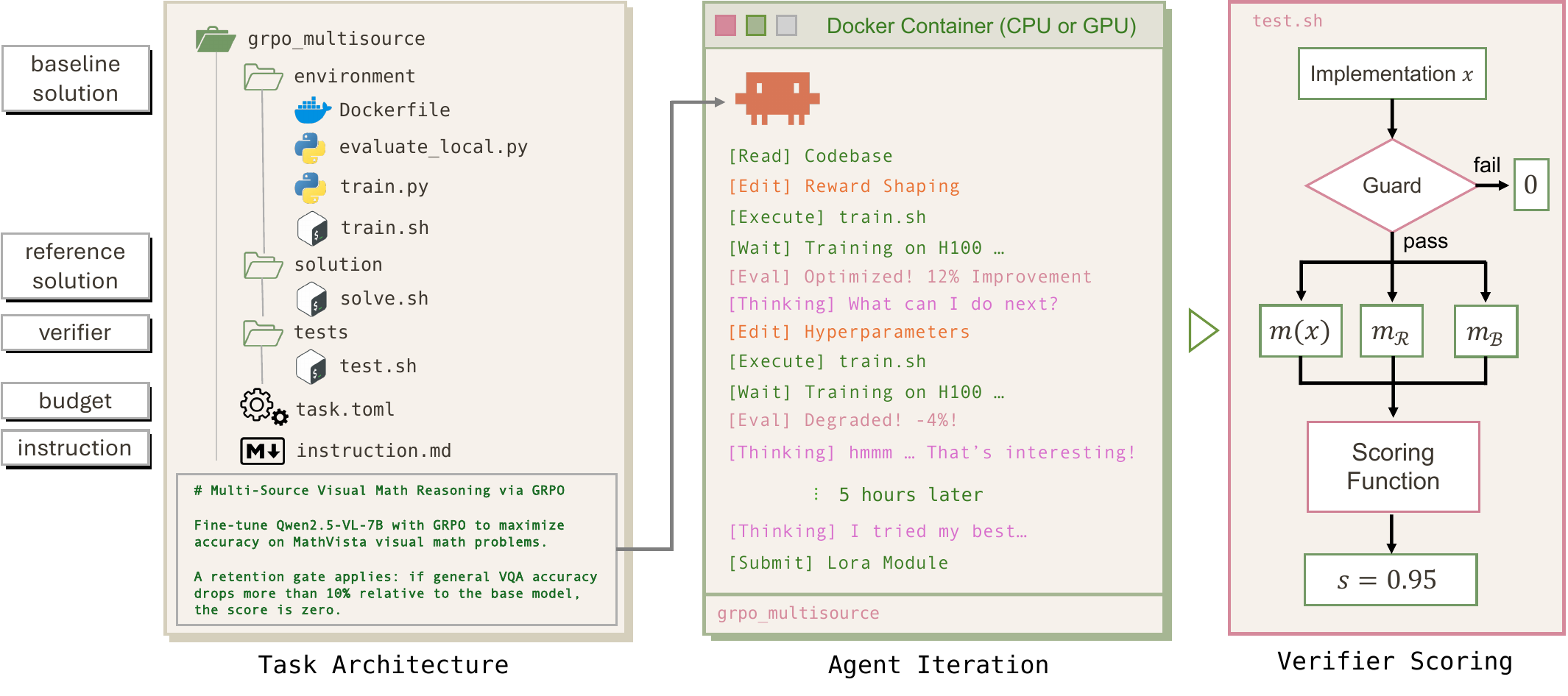}
    \caption{\bench task formulation and evaluation pipeline.}
    \label{fig:pipeline}
\end{figure}

\subsection{Task Formulation}
\label{subsec:task_formulation}

A task in \bench consists of an \emph{instruction}, an
\emph{environment}, a \emph{verifier}, a \emph{reference solution}, and a \emph{wall-clock budget} (Figure \ref{fig:pipeline}). The instruction is a natural-language description of the optimization target. The environment is a containerized sandbox (either CPU or single-GPU depending on the workload) that ships with a codebase containing a working but unoptimized baseline implementation, together with a local evaluation script that the agent may invoke during development. The verifier is the held-out evaluation suite that produces the final score. The reference solution is a human-written implementation that anchors the scoring scale and is never exposed to the agent. The budget bounds the wall-clock time available for the agent to read the codebase, modify it, run it, and iterate.

During evaluation, the agent receives the instruction and environment inside the sandbox, and must produce a modified implementation that the verifier can evaluate within the allotted budget. Tasks are inherently interactive: the agent may freely edit the codebase, execute the implementation, profile its performance, invoke the local evaluation script, inspect intermediate outputs, and iteratively refine its solution. At the end of the episode, the verifier runs the modified implementation on held-out inputs and reports a metric, which is then mapped to a continuous score relative to the reference solution.

\phead{Baselines and references.}
The baselines in \bench tasks are correct but suboptimal, representing where a working but unoptimized first implementation might land. Reference solutions, in contrast, are required to improve the metric by a non-trivial margin (typically at least an order of magnitude on system-optimization tasks, and a clear statistical gain on model-development tasks), so that every task has genuine headroom for an agent to discover.

\phead{Scoring.}
Let \( m(x) \) denote the raw metric value achieved by an implementation \( x \) (e.g., runtime, validation perplexity, throughput, or parameter count), and let \( m_{\mathcal{B}} \) and \( m_{\mathcal{R}} \) denote the metric values attained by the baseline and reference solutions, respectively. \bench{} employs two anchored scoring schemes, both normalized to the interval \([0,1]\) and anchored at the baseline and reference performance levels:

\begin{itemize}[topsep=0pt, itemsep=2pt, leftmargin=*]
  \item \textbf{Log-stretch.} For performance-optimization tasks, where meaningful improvements frequently span orders of magnitude, we adopt a logarithmic scoring scheme:
    \begin{equation}
      s(x) = \mathrm{clip}\!\left( \tfrac{1}{2} \cdot \frac{\log(m_{\mathcal{B}}/m(x))}{\log(m_{\mathcal{B}}/m_{\mathcal{R}})}, \, 0, \, 1 \right)
    \end{equation}
    (with the directional analogue for metrics where higher values are better). A minimum-improvement gate ensures that \( s(x) = 0 \) until the agent exceeds the baseline, yielding \( s = 0 \) at the baseline, \( s = 0.5 \) at the reference, and values approaching \( 1.0 \) as performance nears the practical optimum. This gate prevents submissions that make no meaningful improvement over the baseline from receiving partial credit. We note that both \(m_{\mathcal{B}}\) and \(m_{\mathcal{R}}\) for performance-optimization tasks are \textbf{sandbox-dependent} quantities that have been carefully calibrated for the specific sandbox environments and hardware configurations used throughout this benchmark.

  \item \textbf{Linear.} For tasks with naturally bounded quality metrics, we use linear interpolation between the two anchors:
    \begin{equation}
      s(x) = \mathrm{clip}\!\Big( \frac{m_{\mathcal{B}} - m(x)}{m_{\mathcal{B}} - m_{\mathcal{R}}}, \, 0, \, 1 \Big)
    \end{equation}
    (again with the directional analogue when higher is better). Thus, \( s = 0 \) at the baseline and \( s = 1.0 \) at the reference. 
\end{itemize}

The specific choice of \( m_{\mathcal{B}} \) and \( m_{\mathcal{R}} \), along with any task-specific feasibility gates, is detailed in Appendix~\ref{app:scoring}. Anchored relative scoring serves two important purposes: it enables meaningful aggregation across tasks whose native units are otherwise incommensurable, and unlike binary pass/fail benchmarks, it rewards genuine partial progress. The latter property is particularly crucial at \bench{}'s level of difficulty, where the majority of agent submissions lie between the baseline and the reference solution.

\phead{Wall-clock budget.}
Wall-clock budgets range from 2 hours for the smallest puzzle tasks to 12 hours for end-to-end LLM development tasks. Budgets are chosen to balance two competing goals: preserving realistic development workflows, which often require substantial execution and iteration time, while keeping evaluation costs tractable and reproducible at benchmark scale. Accordingly, some model training tasks are intentionally designed around smaller models and shorter training steps rather than frontier-scale training runs, allowing agents to complete multiple optimization iterations within the time budget. This further challenges agent's ability to allocate time effectively across exploration, execution, and iteration.

\subsection{Benchmark Construction}
\label{subsec:construction}

\phead{Task Collection.}
AutoLab tasks were contributed by senior researchers and engineers. Contributors were asked to draw tasks from real engineering or research problems they had personally encountered, from low-level CUDA and C optimization to end-to-end vision-language model post-training. We deliberately prioritize realism and diversity over difficulty for its own sake: a task earns inclusion because it captures a workflow that practitioners actually undertake, not because it has been calibrated to be artificially hard.

\phead{Quality Control.}
Each task underwent a multi-round audit before inclusion. Inspired by \citet{merrill2026terminalbench}, we audit each task against four criteria tailored to \bench's continuous-scoring, performance-oriented setting: \emph{validity} (a higher score requires a higher-quality implementation, not artifacts of the measurement procedure or weakened correctness checks); \emph{solvability} (the reference solution reliably reaches the target score within the stated budget on the target hardware); \emph{integrity} (the agent cannot pass by hacking the scoring function); and \emph{measurement stability} (the metric variance across repeated runs of the reference is small enough that observed score differences are attributable to the implementation rather than to noise). Each task was reviewed by at least 2 experts independent of the original contributor and a format audit agent; tasks that failed any criterion were either revised or rejected.

\phead{Anti-Reward Hacking.}
Performance benchmarks expose a broader attack surface than patch-style benchmarks. \bench mitigates this risk in five ways. First, the verifier is sealed: the agent is given access to a local evaluation script, but not to the held-out test inputs or reference outputs used for final scoring. Second, ML tasks include a correctness gate that must pass before the optimization metric is recorded, with gate inputs drawn from a distribution disjoint from anything visible during development. Third, we run a dedicated adversarial agent explicitly prompted to discover shortcuts or reward hacks during task construction; any task that can be solved without genuine improvement to the target metric is either patched or removed. Fourth, critical files that should remain immutable are SHA-pinned, and any unauthorized modification immediately results in a zero score. In addition, we continuously analyze agent trajectories across different models during evaluation. When new forms of reward hacking or verifier exploitation are discovered, we patch the corresponding verifier and re-validate affected tasks to maintain benchmark integrity over time.

\begin{figure}[t]
    \centering
    \includegraphics[width=0.95\linewidth]{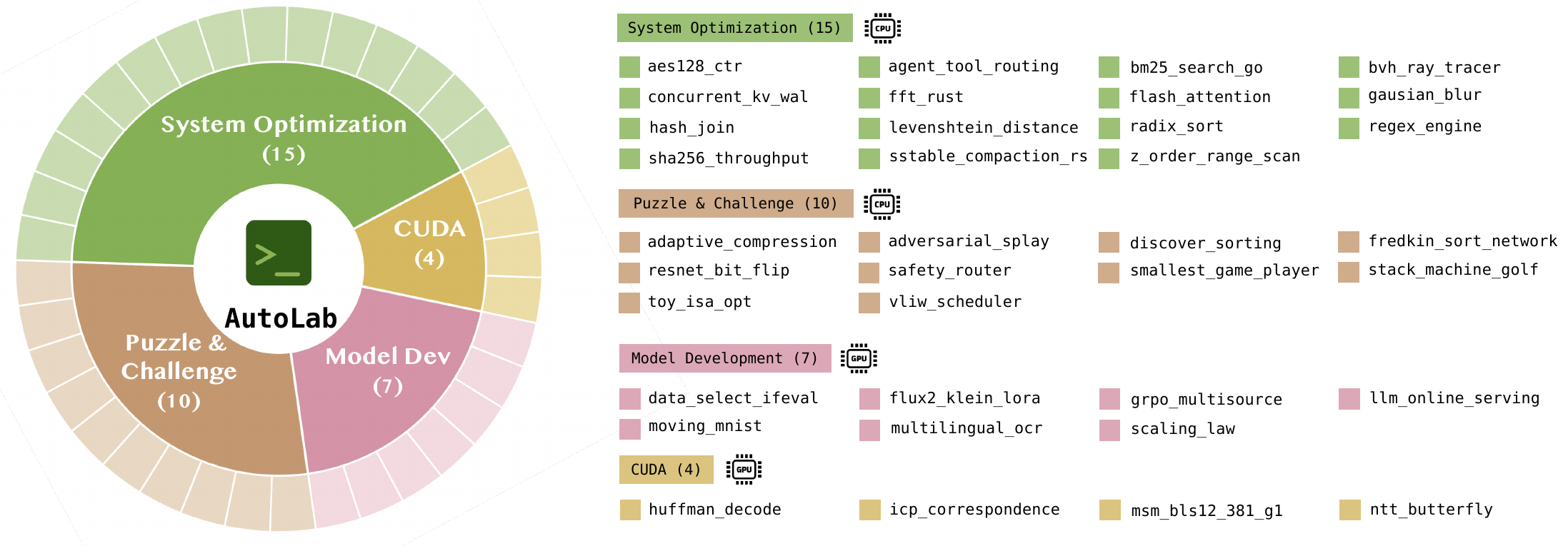}
    \caption{This figure illustrates task distribution of \bench.}
    \label{fig:task_distribution}
\end{figure}

\subsection{Benchmark Composition}
\label{subsec:composition}

The final benchmark comprises 36 tasks across four categories. \emph{Model Development} (7 tasks) covers the full LLM pipeline, including pretraining scaling laws, RL post-training, SFT data selection, parameter-efficient fine-tuning, world-model training, and online serving optimization. \emph{System Optimization} (15 tasks) focuses on low-level performance engineering of systems primitives such as kernels, sorting, hashing, search, compression, regular expressions, and cryptography in C, Rust, Go, and Python. \emph{Puzzle \& Challenge} (10 tasks) consists of algorithmic problems built around a single key insight, including combinatorial reductions, sorting networks, ISA-level scheduling, adversarial constructions, and adaptive coding. \emph{CUDA} (4 tasks) targets GPU kernel optimization for cryptographic primitives, point-cloud registration, and compression. The complete list of tasks appears in Figure~\ref{fig:task_distribution}. Detailed descriptions of each task are provided in Appendix~\ref{app:task-descriptions}.
\section{Benchmark Results}
\label{sec:eval}

\subsection{Experimental Setup}
\label{subsec:exp_setup}

\textbf{Models.}
We evaluate a range of state-of-the-art proprietary and open-weight models in \bench. Proprietary models include \texttt{claude-opus-4.6}~\citep{anthropic2026opus46}, \texttt{gemini-3.1-pro}~\citep{google2026gemini31pro}, \texttt{gpt-5.4}~\citep{openai2026gpt54}, and \texttt{grok-4-20}~\citep{xai2026grok420}. On the open-weight side, we evaluate \texttt{qwen-3.6-plus}~\citep{qwen2026qwen36}, \texttt{deepseek-v4-pro}~\citep{deepseek2026deepseekv4}, \texttt{glm-5}~\citep{zai2026glm5}, \texttt{kimi-k2.6}~\citep{moonshot2026kimik26}, \texttt{hunyuan-3-preview}~\citep{tencent2026hy3}, \texttt{mimo-v2.5-pro}~\citep{xiaomi2026mimov25pro}, and \texttt{minimax-m2.7}~\citep{minimax2026m27}. For ablation analyses, we additionally evaluate older or smaller variants from several of these families: \texttt{kimi-k2.5} \citep{moonshot2025kimik25}, \texttt{minimax-m2.5} \citep{minimax2025m25}, \texttt{mimo-v2-pro} \citep{xiaomi2026mimov2pro}, \texttt{mimo-v2.5} \citep{xiaomi2026mimov25}, \texttt{deepseek-v4-flash} \citep{deepseek2026deepseekv4}, and \texttt{qwen-3.5-plus} \citep{qwen2026qwen35}. We do not test small open-weight models ($<$200B parameters) due to the difficulty of the benchmark. The specific API provider used to access each model is listed in Table \ref{tab:model_api_providers} in Appendix \ref{app:more_exp_setups} for better reproducibility.

\phead{Evaluation metrics.} 
We evaluate every (model, task) pair with \textbf{three} independent rollouts and report three complementary metrics. 
\textbf{Avg@3} averages the per-run score across the three trials, capturing typical performance. 
\textbf{Best@3} takes the maximum of the three trials, reflecting an agent's ceiling. 
\textbf{Dominance} measures a model's head-to-head win rate against all other models using Avg@3 scores. Formally, let \(\mathcal{M}\) be the set of evaluated models and \(\mathcal{T}\) the set of tasks, and let \(s_{m,t}\) denote model \(m\)'s Avg@3 score on task \(t\). Then
\begin{equation}
\mathrm{Dominance}(m) 
= \frac{1}{|\mathcal{T}| \cdot (|\mathcal{M}|-1)} 
\sum_{t \in \mathcal{T}} 
\sum_{\substack{o \in \mathcal{M} \\ o \neq m}} 
\Bigl( \mathbf{1}[s_{m,t} > s_{o,t}] + \tfrac{1}{2}\,\mathbf{1}[s_{m,t} = s_{o,t}] \Bigr).
\end{equation}
Thus, \(\mathrm{Dominance}(m) \in [0, 1]\), where \(1\) means the model strictly outperforms every other model on every task and \(0.5\) corresponds to average performance across models. 
This metric provides a robust, tournament-style view that is largely insensitive to hardware variance and differences in per-task reward design, while being less sensitive to a small number of high-leverage tasks.

\phead{Implementation Details.} Following Terminal-Bench \citep{merrill2026terminalbench}, we use the \textsc{Harbor} framework \citep{Harbor_Framework} as the unified evaluation harness, and the \texttt{terminus-2} agent by default across all models. While specialized harnesses may further improve performance, we leave such optimizations to future work, and provide an early pilot comparison with two alternative agent harnesses, \texttt{pi-mono} \citep{pi_mono} and an optimized \texttt{mini-swe-agent} \citep{yang2024sweagent}, in Section~\ref{app:harness}. CPU-only tasks run inside a local Docker sandbox, and GPU tasks run on Modal\footnote{\url{https://modal.com}} cloud sandboxes provisioned with H100 and L40S GPUs. The local CPU sandbox runs on a workstation with an AMD Ryzen~9 9950X (16~cores / 32 threads) and 64\,GB of RAM, and per-task CPU and memory caps are enforced via the task's metadata. In total, the evaluation of \bench consumed \textbf{2{,}544 wall-clock hours} and \textbf{8.60 billion tokens}.

\subsection{Main Results}
\label{subsec:main_results}

\begin{table*}[t]
\centering
\footnotesize
\setlength{\tabcolsep}{3pt}
\renewcommand{\arraystretch}{1.30}
\setlength{\aboverulesep}{1pt}
\setlength{\belowrulesep}{1pt}
\resizebox{\textwidth}{!}{%
\begin{tabular}{lccccccccccccccc}
\toprule
\textbf{Model} & \multicolumn{3}{c}{\textbf{\shortstack{Overall\\\,}}} & \multicolumn{3}{c}{\textbf{\shortstack{CUDA\\\,}}} & \multicolumn{3}{c}{\textbf{\shortstack{Model\\Development}}} & \multicolumn{3}{c}{\textbf{\shortstack{Puzzle \&\\Challenge}}} & \multicolumn{3}{c}{\textbf{\shortstack{System\\Optimization}}} \\
\cmidrule(lr){2-4} \cmidrule(lr){5-7} \cmidrule(lr){8-10} \cmidrule(lr){11-13} \cmidrule(lr){14-16}
 & Avg\,$\uparrow$ & Best\,$\uparrow$ & Dom.\,$\uparrow$ & Avg\,$\uparrow$ & Best\,$\uparrow$ & Dom.\,$\uparrow$ & Avg\,$\uparrow$ & Best\,$\uparrow$ & Dom.\,$\uparrow$ & Avg\,$\uparrow$ & Best\,$\uparrow$ & Dom.\,$\uparrow$ & Avg\,$\uparrow$ & Best\,$\uparrow$ & Dom.\,$\uparrow$ \\
\specialrule{0.9pt}{0pt}{0pt}
\addlinespace[2pt]
\multicolumn{16}{l}{\textit{\textbf{Main Set}}} \\
\addlinespace[2pt]
\specialrule{0.9pt}{0pt}{0pt}
\raisebox{-0.30ex}{\includegraphics[height=2.4ex]{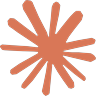}}~\texttt{claude-opus-4.6} & \cellcolor[HTML]{DFE9D5} \textbf{0.68} & \cellcolor[HTML]{D1DFC3} \textbf{0.76} & \cellcolor[HTML]{B1C89A} \textbf{0.93} & \cellcolor[HTML]{FAF0F2} \textbf{0.38} & \cellcolor[HTML]{F9FAF7} \textbf{0.54} & \cellcolor[HTML]{A9C390} \textbf{0.97} & \cellcolor[HTML]{E8EFE1} \textbf{0.63} & \cellcolor[HTML]{D7E3CB} \textbf{0.72} & \cellcolor[HTML]{BCD0A8} \textbf{0.87} & \cellcolor[HTML]{C1D3AE} \textbf{0.85} & \cellcolor[HTML]{B1C89A} \textbf{0.93} & \cellcolor[HTML]{ADC594} \textbf{0.95} & \cellcolor[HTML]{E1EAD8} \textbf{0.67} & \cellcolor[HTML]{D9E4CD} \textbf{0.71} & \cellcolor[HTML]{B1C899} \textbf{0.93} \\
\noalign{\vskip1pt}\arrayrulecolor{black!18}\specialrule{0.3pt}{0pt}{0pt}\arrayrulecolor{black}\noalign{\vskip1pt}
\raisebox{-0.30ex}{\includegraphics[height=2.4ex]{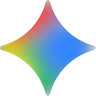}}~\texttt{gemini-3.1-pro} & \underline{0.50} & \cellcolor[HTML]{EFF4EA} 0.59 & \cellcolor[HTML]{E9F0E3} \underline{0.62} & \cellcolor[HTML]{F2DBE1} 0.22 & \cellcolor[HTML]{F3DEE3} 0.25 & \cellcolor[HTML]{F8FAF6} 0.54 & \cellcolor[HTML]{F8ECEF} 0.36 & 0.48 & \cellcolor[HTML]{FDF9FA} 0.46 & \cellcolor[HTML]{D8E4CC} \underline{0.72} & \cellcolor[HTML]{C7D8B6} \underline{0.81} & \cellcolor[HTML]{D0DEC2} \underline{0.76} & 0.49 & \cellcolor[HTML]{F0F4EB} 0.59 & \cellcolor[HTML]{E8EFE1} 0.63 \\
\noalign{\vskip1pt}\arrayrulecolor{black!18}\specialrule{0.3pt}{0pt}{0pt}\arrayrulecolor{black}\noalign{\vskip1pt}
\raisebox{-0.30ex}{\includegraphics[height=2.4ex]{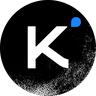}}~\texttt{kimi-k2.6} & \cellcolor[HTML]{FDFAFB} 0.46 & \cellcolor[HTML]{ECF2E7} \underline{0.60} & \cellcolor[HTML]{E9F0E3} 0.62 & \cellcolor[HTML]{F3DEE3} \underline{0.25} & \cellcolor[HTML]{FCF5F7} \underline{0.43} & \cellcolor[HTML]{D6E3CA} 0.72 & \cellcolor[HTML]{FBF4F6} 0.42 & \cellcolor[HTML]{FAFBF8} 0.53 & 0.49 & \cellcolor[HTML]{FEFCFD} 0.48 & \cellcolor[HTML]{E0E9D7} 0.67 & \cellcolor[HTML]{FBFCFA} 0.52 & \cellcolor[HTML]{FBFCF9} \underline{0.52} & \cellcolor[HTML]{E6EDDE} \underline{0.64} & \cellcolor[HTML]{D7E3CB} \underline{0.72} \\
\noalign{\vskip1pt}\arrayrulecolor{black!18}\specialrule{0.3pt}{0pt}{0pt}\arrayrulecolor{black}\noalign{\vskip1pt}
\raisebox{-0.30ex}{\includegraphics[height=2.4ex]{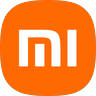}}~\texttt{mimo-v2.5-pro} & \cellcolor[HTML]{FDF9FA} 0.45 & \cellcolor[HTML]{F1F5ED} 0.58 & \cellcolor[HTML]{F9FBF7} 0.53 & \cellcolor[HTML]{F0D5DB} 0.18 & \cellcolor[HTML]{F6E8EB} 0.32 & \cellcolor[HTML]{FEFCFC} 0.47 & \cellcolor[HTML]{FEFCFC} \underline{0.47} & \cellcolor[HTML]{EBF1E5} 0.61 & \cellcolor[HTML]{EAF1E4} \underline{0.61} & \cellcolor[HTML]{E7EEDF} 0.64 & \cellcolor[HTML]{CDDCBE} 0.78 & \cellcolor[HTML]{EAF0E4} 0.61 & \cellcolor[HTML]{FAF1F3} 0.39 & 0.49 & \cellcolor[HTML]{FDFAFB} 0.46 \\
\noalign{\vskip1pt}\arrayrulecolor{black!18}\specialrule{0.3pt}{0pt}{0pt}\arrayrulecolor{black}\noalign{\vskip1pt}
\raisebox{-0.30ex}{\includegraphics[height=2.4ex]{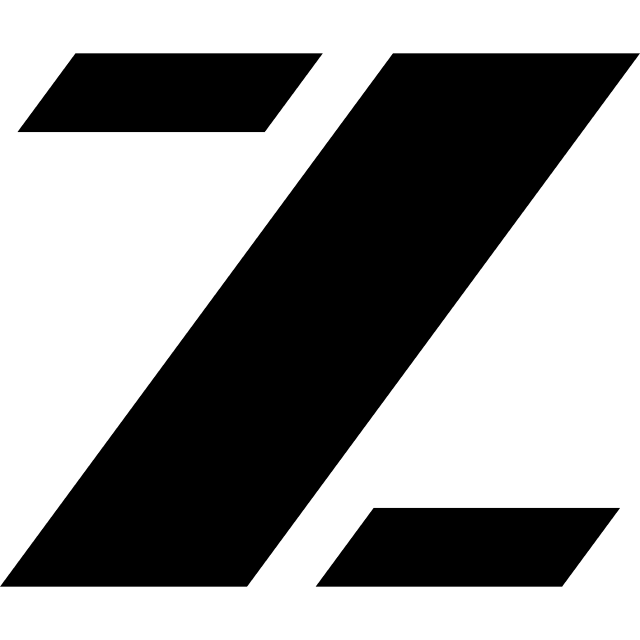}}~\texttt{glm-5} & \cellcolor[HTML]{FCF7F8} 0.43 & \cellcolor[HTML]{F6F8F3} 0.55 & \cellcolor[HTML]{F2F6EE} 0.57 & \cellcolor[HTML]{F3DEE3} 0.24 & \cellcolor[HTML]{FAF1F3} 0.39 & \cellcolor[HTML]{D2DFC4} \underline{0.75} & \cellcolor[HTML]{FDF8F9} 0.45 & \cellcolor[HTML]{ECF2E7} 0.60 & \cellcolor[HTML]{F0F4EB} 0.59 & 0.49 & \cellcolor[HTML]{EBF1E5} 0.61 & \cellcolor[HTML]{FDF9FA} 0.46 & \cellcolor[HTML]{FCF7F8} 0.44 & \cellcolor[HTML]{F9FBF7} 0.53 & \cellcolor[HTML]{EEF3E9} 0.59 \\
\noalign{\vskip1pt}\arrayrulecolor{black!18}\specialrule{0.3pt}{0pt}{0pt}\arrayrulecolor{black}\noalign{\vskip1pt}
\raisebox{-0.30ex}{\includegraphics[height=2.4ex]{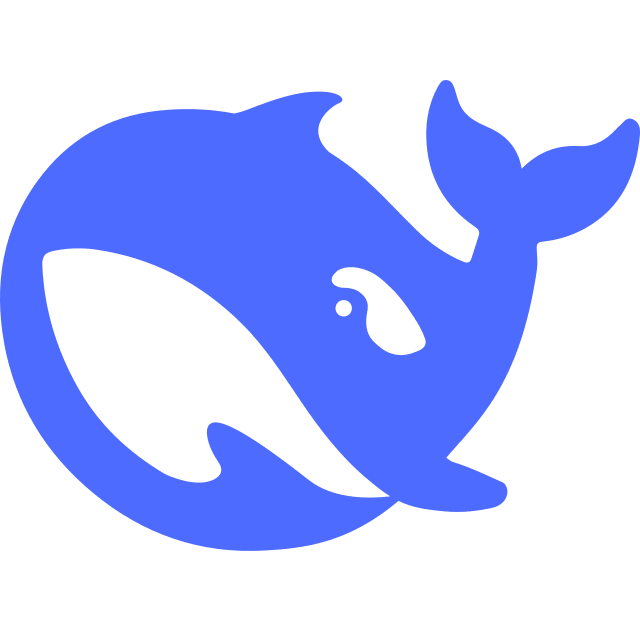}}~\texttt{deepseek-v4-pro} & \cellcolor[HTML]{F9EFF2} 0.38 & 0.51 & \cellcolor[HTML]{FDFBFB} 0.47 & \cellcolor[HTML]{E7BEC8} 0.00 & \cellcolor[HTML]{E7BEC8} 0.00 & \cellcolor[HTML]{F3DFE4} 0.25 & \cellcolor[HTML]{F8ECEF} 0.35 & \cellcolor[HTML]{F4F8F1} 0.56 & \cellcolor[HTML]{FEFBFC} 0.47 & \cellcolor[HTML]{FCF6F8} 0.43 & \cellcolor[HTML]{E1EAD8} 0.67 & \cellcolor[HTML]{FAF1F3} 0.39 & \cellcolor[HTML]{FDF9FA} 0.46 & 0.52 & \cellcolor[HTML]{F1F5ED} 0.58 \\
\noalign{\vskip1pt}\arrayrulecolor{black!18}\specialrule{0.3pt}{0pt}{0pt}\arrayrulecolor{black}\noalign{\vskip1pt}
\raisebox{-0.30ex}{\includegraphics[height=2.4ex]{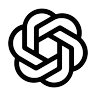}}~\texttt{gpt-5.4} & \cellcolor[HTML]{F8EDF0} 0.36 & \cellcolor[HTML]{FAFCF9} 0.53 & \cellcolor[HTML]{FAF1F3} 0.39 & \cellcolor[HTML]{EED0D7} 0.14 & \cellcolor[HTML]{F1D8DE} 0.20 & \cellcolor[HTML]{FAF0F3} 0.39 & \cellcolor[HTML]{F8ECEF} 0.35 & \cellcolor[HTML]{F7F9F4} 0.55 & \cellcolor[HTML]{FBF4F6} 0.41 & \cellcolor[HTML]{FDF8F9} 0.45 & \cellcolor[HTML]{DEE8D4} 0.68 & \cellcolor[HTML]{FDF9FA} 0.45 & \cellcolor[HTML]{F9EEF1} 0.37 & 0.50 & \cellcolor[HTML]{F8EBEE} 0.35 \\
\noalign{\vskip1pt}\arrayrulecolor{black!18}\specialrule{0.3pt}{0pt}{0pt}\arrayrulecolor{black}\noalign{\vskip1pt}
\raisebox{-0.30ex}{\includegraphics[height=2.4ex]{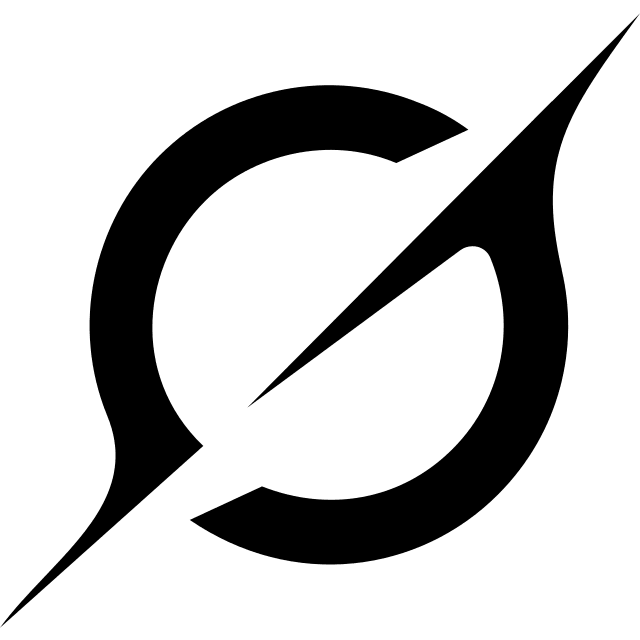}}~\texttt{grok-4-20} & \cellcolor[HTML]{F8ECEF} 0.35 & \cellcolor[HTML]{FCF7F9} 0.44 & \cellcolor[HTML]{FBF5F7} 0.42 & \cellcolor[HTML]{F0D5DC} 0.18 & \cellcolor[HTML]{F5E2E7} 0.28 & \cellcolor[HTML]{E8EFE2} 0.62 & \cellcolor[HTML]{EECFD7} 0.13 & \cellcolor[HTML]{F1D9DF} 0.21 & \cellcolor[HTML]{F8EBEE} 0.34 & \cellcolor[HTML]{F9FBF8} 0.53 & \cellcolor[HTML]{DEE8D4} 0.68 & \cellcolor[HTML]{FDFAFB} 0.47 & \cellcolor[HTML]{F9EFF2} 0.38 & \cellcolor[HTML]{FCF6F8} 0.43 & \cellcolor[HTML]{F9EFF2} 0.38 \\
\noalign{\vskip1pt}\arrayrulecolor{black!18}\specialrule{0.3pt}{0pt}{0pt}\arrayrulecolor{black}\noalign{\vskip1pt}
\raisebox{-0.30ex}{\includegraphics[height=2.4ex]{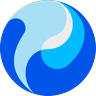}}~\texttt{hunyuan-3-preview} & \cellcolor[HTML]{F6E6EA} 0.31 & \cellcolor[HTML]{FCF8F9} 0.45 & \cellcolor[HTML]{F7EAED} 0.34 & \cellcolor[HTML]{EBC7D0} 0.07 & \cellcolor[HTML]{F1D9DF} 0.21 & \cellcolor[HTML]{F6E5E9} 0.30 & \cellcolor[HTML]{F4E1E6} 0.27 & \cellcolor[HTML]{F9EFF2} 0.38 & \cellcolor[HTML]{F3DEE3} 0.24 & \cellcolor[HTML]{F8EBEE} 0.35 & \cellcolor[HTML]{F6F9F3} 0.55 & \cellcolor[HTML]{F8EDF0} 0.36 & \cellcolor[HTML]{F9EDF0} 0.36 & \cellcolor[HTML]{FEFBFC} 0.47 & \cellcolor[HTML]{F9EFF1} 0.37 \\
\noalign{\vskip1pt}\arrayrulecolor{black!18}\specialrule{0.3pt}{0pt}{0pt}\arrayrulecolor{black}\noalign{\vskip1pt}
\raisebox{-0.30ex}{\includegraphics[height=2.4ex]{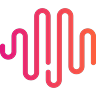}}~\texttt{minimax-m2.7} & \cellcolor[HTML]{F4E1E6} 0.27 & \cellcolor[HTML]{FCF6F7} 0.43 & \cellcolor[HTML]{F5E3E7} 0.28 & \cellcolor[HTML]{EDCED6} 0.12 & \cellcolor[HTML]{F0D7DD} 0.19 & \cellcolor[HTML]{F8ECEE} 0.35 & \cellcolor[HTML]{FBF5F7} 0.42 & \cellcolor[HTML]{E7EEDF} \underline{0.64} & \cellcolor[HTML]{F6F9F3} 0.55 & \cellcolor[HTML]{F4E0E5} 0.26 & \cellcolor[HTML]{FCF7F8} 0.44 & \cellcolor[HTML]{F2DCE1} 0.23 & \cellcolor[HTML]{F3DEE3} 0.25 & \cellcolor[HTML]{FAF1F3} 0.39 & \cellcolor[HTML]{F0D5DC} 0.18 \\
\noalign{\vskip1pt}\arrayrulecolor{black!18}\specialrule{0.3pt}{0pt}{0pt}\arrayrulecolor{black}\noalign{\vskip1pt}
\raisebox{-0.30ex}{\includegraphics[height=2.4ex]{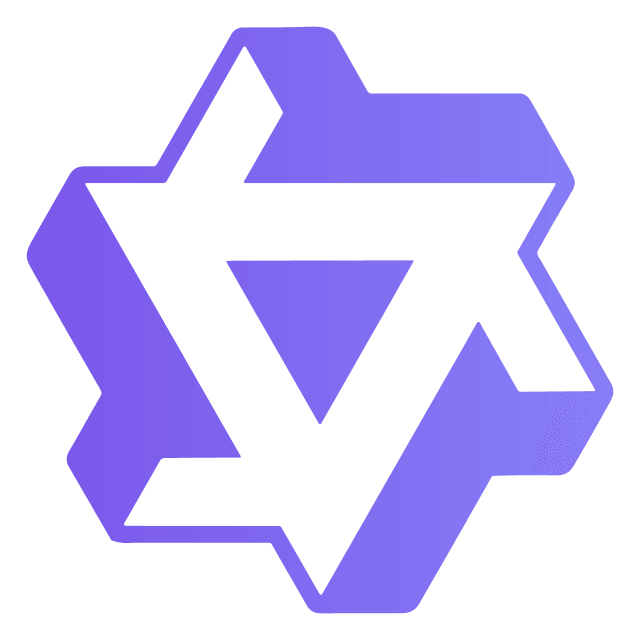}}~\texttt{qwen-3.6-plus} & \cellcolor[HTML]{F4E1E5} 0.27 & \cellcolor[HTML]{FAF0F3} 0.39 & \cellcolor[HTML]{F6E7EB} 0.32 & \cellcolor[HTML]{E7BEC8} 0.00 & \cellcolor[HTML]{E7BEC8} 0.00 & \cellcolor[HTML]{EDCED6} 0.12 & \cellcolor[HTML]{FAF1F3} 0.39 & \cellcolor[HTML]{EBF1E5} 0.61 & \cellcolor[HTML]{FDF9FA} 0.46 & \cellcolor[HTML]{F4E0E4} 0.26 & \cellcolor[HTML]{F9EFF2} 0.38 & \cellcolor[HTML]{F6E5E9} 0.30 & \cellcolor[HTML]{F5E4E8} 0.29 & \cellcolor[HTML]{FAF1F3} 0.39 & \cellcolor[HTML]{F6E6EA} 0.31 \\
\specialrule{0.9pt}{0pt}{0pt}
\addlinespace[2pt]
\multicolumn{16}{l}{\textit{\textbf{Ablation Set}}} \\
\addlinespace[2pt]
\specialrule{0.9pt}{0pt}{0pt}
\raisebox{-0.30ex}{\includegraphics[height=2.4ex]{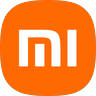}}~\texttt{mimo-v2.5} & \cellcolor[HTML]{F7EAED} 0.34 & 0.51 & \multicolumn{1}{c}{--} & \cellcolor[HTML]{EFD2D9} 0.16 & \cellcolor[HTML]{F5E3E7} 0.28 & \multicolumn{1}{c}{--} & \cellcolor[HTML]{F9EDF0} 0.36 & 0.52 & \multicolumn{1}{c}{--} & \cellcolor[HTML]{FAF2F4} 0.40 & \cellcolor[HTML]{DCE7D2} 0.69 & \multicolumn{1}{c}{--} & \cellcolor[HTML]{F7EAED} 0.34 & \cellcolor[HTML]{FDF9FA} 0.45 & \multicolumn{1}{c}{--} \\
\noalign{\vskip1pt}\arrayrulecolor{black!18}\specialrule{0.3pt}{0pt}{0pt}\arrayrulecolor{black}\noalign{\vskip1pt}
\raisebox{-0.30ex}{\includegraphics[height=2.4ex]{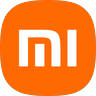}}~\texttt{mimo-v2-pro} & \cellcolor[HTML]{F9EDF0} 0.37 & 0.51 & \multicolumn{1}{c}{--} & \cellcolor[HTML]{F4E1E6} 0.27 & \cellcolor[HTML]{FCF7F8} 0.44 & \multicolumn{1}{c}{--} & \cellcolor[HTML]{F8ECEF} 0.35 & \cellcolor[HTML]{FDF8F9} 0.45 & \multicolumn{1}{c}{--} & \cellcolor[HTML]{FCF8F9} 0.45 & \cellcolor[HTML]{EEF3E9} 0.59 & \multicolumn{1}{c}{--} & \cellcolor[HTML]{F8EBEE} 0.34 & 0.50 & \multicolumn{1}{c}{--} \\
\noalign{\vskip1pt}\arrayrulecolor{black!18}\specialrule{0.3pt}{0pt}{0pt}\arrayrulecolor{black}\noalign{\vskip1pt}
\raisebox{-0.30ex}{\includegraphics[height=2.4ex]{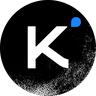}}~\texttt{kimi-k2.5} & \cellcolor[HTML]{FAF2F4} 0.40 & \cellcolor[HTML]{F8FAF5} 0.54 & \multicolumn{1}{c}{--} & \cellcolor[HTML]{F5E3E7} 0.28 & \cellcolor[HTML]{FBF4F6} 0.42 & \multicolumn{1}{c}{--} & \cellcolor[HTML]{FDFAFB} 0.46 & \cellcolor[HTML]{F2F6EE} 0.57 & \multicolumn{1}{c}{--} & \cellcolor[HTML]{FDF9FA} 0.45 & \cellcolor[HTML]{E9F0E3} 0.62 & \multicolumn{1}{c}{--} & \cellcolor[HTML]{F9EEF1} 0.37 & 0.51 & \multicolumn{1}{c}{--} \\
\noalign{\vskip1pt}\arrayrulecolor{black!18}\specialrule{0.3pt}{0pt}{0pt}\arrayrulecolor{black}\noalign{\vskip1pt}
\raisebox{-0.30ex}{\includegraphics[height=2.4ex]{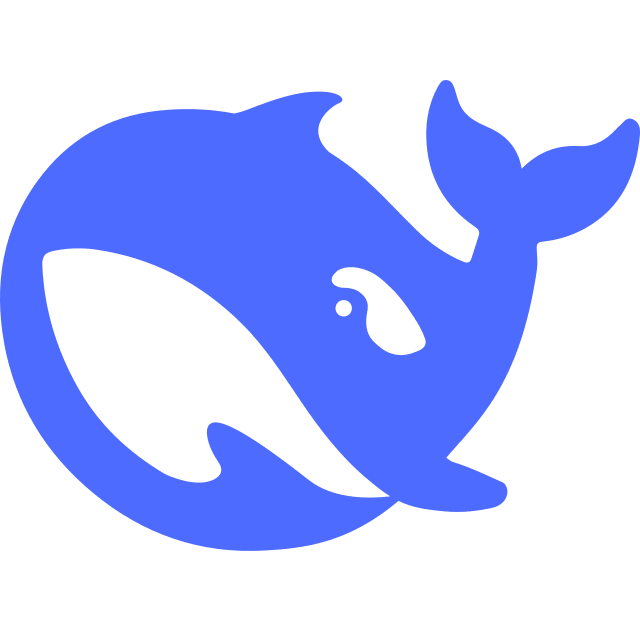}}~\texttt{deepseek-v4-flash} & \cellcolor[HTML]{F9EEF1} 0.37 & \cellcolor[HTML]{FAFBF8} 0.53 & \multicolumn{1}{c}{--} & \cellcolor[HTML]{F3DDE2} 0.24 & \cellcolor[HTML]{FBF3F5} 0.41 & \multicolumn{1}{c}{--} & \cellcolor[HTML]{F6E7EB} 0.31 & \cellcolor[HTML]{FCF8F9} 0.45 & \multicolumn{1}{c}{--} & 0.49 & \cellcolor[HTML]{D7E3CB} 0.72 & \multicolumn{1}{c}{--} & \cellcolor[HTML]{F8ECEF} 0.35 & \cellcolor[HTML]{FEFBFC} 0.47 & \multicolumn{1}{c}{--} \\
\noalign{\vskip1pt}\arrayrulecolor{black!18}\specialrule{0.3pt}{0pt}{0pt}\arrayrulecolor{black}\noalign{\vskip1pt}
\raisebox{-0.30ex}{\includegraphics[height=2.4ex]{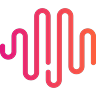}}~\texttt{minimax-m2.5} & \cellcolor[HTML]{F3DEE3} 0.24 & \cellcolor[HTML]{F8EDEF} 0.36 & \multicolumn{1}{c}{--} & \cellcolor[HTML]{EBC7D0} 0.07 & \cellcolor[HTML]{EDCED5} 0.12 & \multicolumn{1}{c}{--} & \cellcolor[HTML]{F8ECEF} 0.36 & \cellcolor[HTML]{FBF4F6} 0.41 & \multicolumn{1}{c}{--} & \cellcolor[HTML]{F4E2E6} 0.27 & \cellcolor[HTML]{FCF6F8} 0.43 & \multicolumn{1}{c}{--} & \cellcolor[HTML]{F2DAE0} 0.22 & \cellcolor[HTML]{F8EBEE} 0.34 & \multicolumn{1}{c}{--} \\
\noalign{\vskip1pt}\arrayrulecolor{black!18}\specialrule{0.3pt}{0pt}{0pt}\arrayrulecolor{black}\noalign{\vskip1pt}
\raisebox{-0.30ex}{\includegraphics[height=2.4ex]{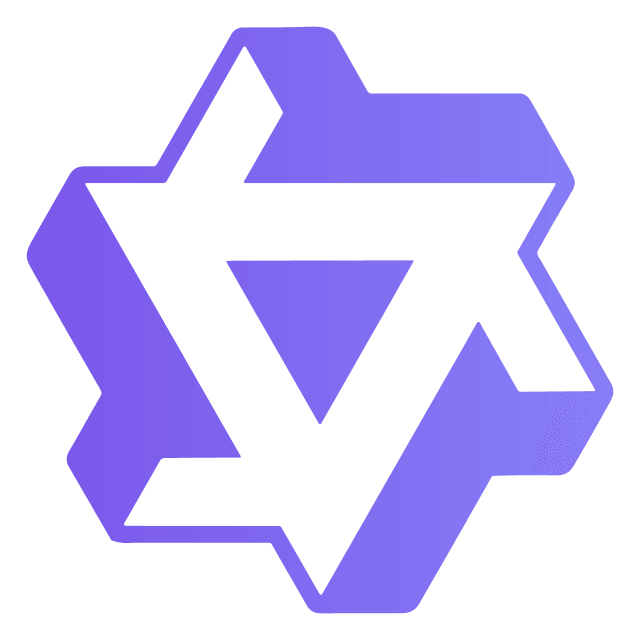}}~\texttt{qwen-3.5-plus} & \cellcolor[HTML]{F9EEF1} 0.37 & 0.50 & \multicolumn{1}{c}{--} & \cellcolor[HTML]{F3DEE3} 0.25 & \cellcolor[HTML]{F7E8EB} 0.32 & \multicolumn{1}{c}{--} & \cellcolor[HTML]{FBF5F6} 0.42 & \cellcolor[HTML]{F5F8F3} 0.55 & \multicolumn{1}{c}{--} & \cellcolor[HTML]{F9EEF1} 0.37 & \cellcolor[HTML]{F2F6EF} 0.57 & \multicolumn{1}{c}{--} & \cellcolor[HTML]{FAF0F2} 0.39 & \cellcolor[HTML]{FEFCFC} 0.48 & \multicolumn{1}{c}{--} \\
\bottomrule
\end{tabular}%
}
\caption{This table shows the per-category results on the \bench benchmark. For each sub-domain we report Avg@3, Best@3, and Dominance. Per-column best scores are shown in \textbf{bold} and runner-up scores are \underline{underlined} (computed on the main set only). Main-set rows are ordered by overall Avg@3.}
\vspace{-1em}
\label{tab:main}
\end{table*}

\phead{Overall Performance.}
Table~\ref{tab:main} presents the Avg@3, Best@3, and Dominance scores for all evaluated models across the 36 tasks. \texttt{claude-opus-4.6} leads the benchmark by a substantial margin, achieving an Avg@3 of $0.68$ and a Dominance score of $0.93$. The performance gap to the second-place model, \texttt{gemini-3.1-pro} (Avg@3 of $0.50$), is large and highlights a clear separation among frontier models on long-horizon iterative improvement tasks.
Other proprietary frontier models such as \texttt{gpt-5.4} and \texttt{grok-4-20} rank in the lower half of the leaderboard. We attribute this primarily to their tendency toward premature termination (see Section~\ref{sec:analysis}) rather than insufficient raw capability. Among open-weight models, \texttt{kimi-k2.6} ($0.46$), \texttt{mimo-v2.5-pro} ($0.45$), and \texttt{glm-5} ($0.43$) form a strong and tight cluster. Notably, smaller models such as \texttt{mimo-v2.5} and \texttt{deepseek-v4-flash} (both under 400B parameters) remain highly competitive. In particular, \texttt{deepseek-v4-flash} ($0.37$) performs on par with the much larger \texttt{deepseek-v4-pro} ($0.38$) overall, and even surpasses it on CUDA kernel optimization and algorithmic puzzle tasks. Detailed per-task results are provided in Appendix~\ref{app:detailed_exp_results}.

\phead{Performance by Category.}
The breakdown across the four task categories (Model Development, System Optimization, CUDA, and Puzzle \& Challenge) reveals distinct model strengths. \texttt{claude-opus-4.6} leads in all four categories by a wide margin, with its largest advantage appearing on CUDA tasks, where most other models score near zero. \texttt{gemini-3.1-pro} performs best on puzzle tasks but lags significantly on CUDA and model development tasks, consistent with its relatively short rollouts (median of 12 steps versus 57 steps for \texttt{claude-opus-4.6}). Open-weight models show their strongest results on system optimization tasks, while CUDA kernel optimization remains a notable weakness across this group.

\begin{figure}[t]
\centering
\includegraphics[width=0.9\textwidth]{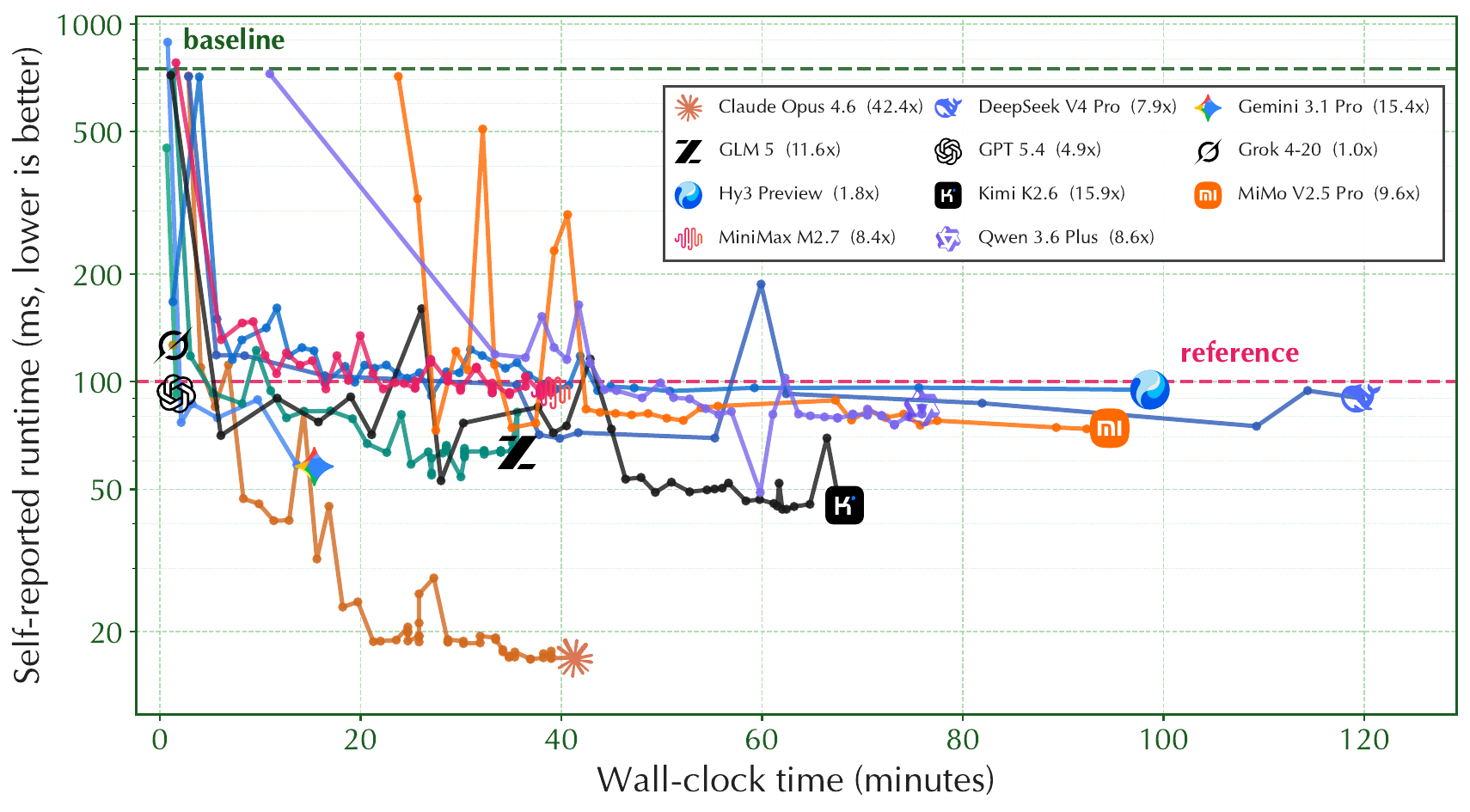}
\caption{Self-reported runtime of each model’s best \texttt{flash\_attention} rollout as a function of wall-clock time. Lower is better. Dashed lines indicate the task baseline ($750$ ms) and the reference solution ($100$ ms). Numbers in the legend report the end-to-end speedup achieved relative to the task baseline.}
\label{fig:flashattn-improvement}
\end{figure}

\phead{Case Study: Flash Attention Optimization.}
To illustrate divergent optimization behaviors, we analyze the best rollout of each model on the \texttt{flash\_attention} task, a two-hour CPU kernel optimization challenge requiring a tiled attention kernel in single-threaded C. All models begin from the same baseline runtime of approximately $750$\,ms, but their trajectories diverge sharply (Figure~\ref{fig:flashattn-improvement}).
\texttt{claude-opus-4.6} steadily reduces runtime to $18$\,ms through 44 feedback-driven iterations over roughly 40 minutes, achieving a $42.4\times$ speedup and surpassing the reference solution ($100$\,ms). In contrast, several strong models plateau near or above the reference: \texttt{kimi-k2.6}, \texttt{gemini-3.1-pro}, and \texttt{glm-5} reach $50$--$80$\,ms but fail to improve further.

Reasoning-heavy models such as \texttt{mimo-v2.5-pro} and \texttt{deepseek-v4-pro} exhibit a distinct failure mode. They spend the majority of their budget on prolonged per-step thinking rather than command execution, which severely delays the first benchmark and limits the total number of edit-and-rerun iterations. Consequently, \texttt{deepseek-v4-pro}’s final submitted solution is not its best result within the trajectory, as it times out before fully exploiting promising directions.
Additionally, \texttt{qwen-3.6-plus} briefly reached a strong intermediate result (better than its final submission) but discarded it after incorrectly judging the solution as illegal. At the low end, \texttt{grok-4-20} and \texttt{gpt-5.4} show minimal progress, with \texttt{grok-4-20} running the evaluation script only once before early termination.
This case study highlights a recurring pattern on \bench{}: high final performance demands not only strong initial coding ability, but sustained, measurement-driven iteration coupled with effective time awareness and self-verification.
\section{Analysis}
\label{sec:analysis}

\subsection{Cost Analysis}
\label{sec:cost-analysis}

\begin{figure}[t]
    \centering
    \includegraphics[width=\textwidth]{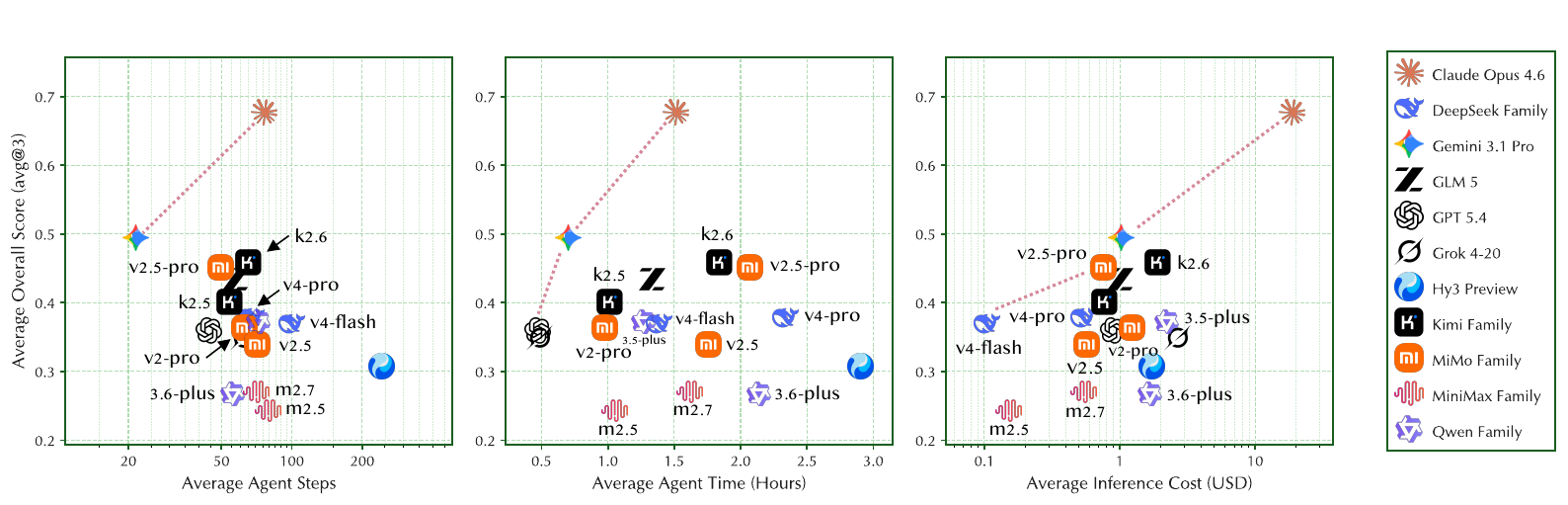}
    \caption{Relationship between models' Avg@3 and three measures of resource utilization. 
    Left: average agent steps. 
    Middle: average agent runtime in hours. 
    Right: average inference cost in USD.}
    \label{fig:score_vs_resource}
\end{figure}

Figure~\ref{fig:score_vs_resource} examines the relationship between models' average overall score and three measures of resource utilization: \emph{average agent steps} (left panel), \emph{average agent runtime} in hours (middle panel), and \emph{average inference cost} in USD (right panel). A clear positive correlation is evident in the left and middle panels between average overall score and both the number of agent steps and total wall-clock runtime. 
\texttt{claude-opus-4.6} stands out prominently as an outlier, requiring substantially more steps than most other models while achieving the highest average score. 
In contrast, the short-horizon termination behavior discussed earlier is clearly visible in the middle panel (agent runtime): models such as \texttt{gpt-5.4} and \texttt{grok-4-20} cluster at markedly lower runtimes, indicating that they terminate trials prematurely rather than continuing to iterate. This early stopping directly limits their optimization potential and largely explains why several proprietary frontier models rank low in the benchmark.

On the inference-cost dimension (right panel), higher performance generally incurs higher costs. Several open-weight models, particularly, \texttt{deepseek-v4-flash} and \texttt{mimo-v2.5-pro}, achieve competitive performance at substantially lower inference costs, highlighting promising avenues for cost-efficient model and agent design.

\begin{figure}[t]
  \centering
  \includegraphics[width=\linewidth]{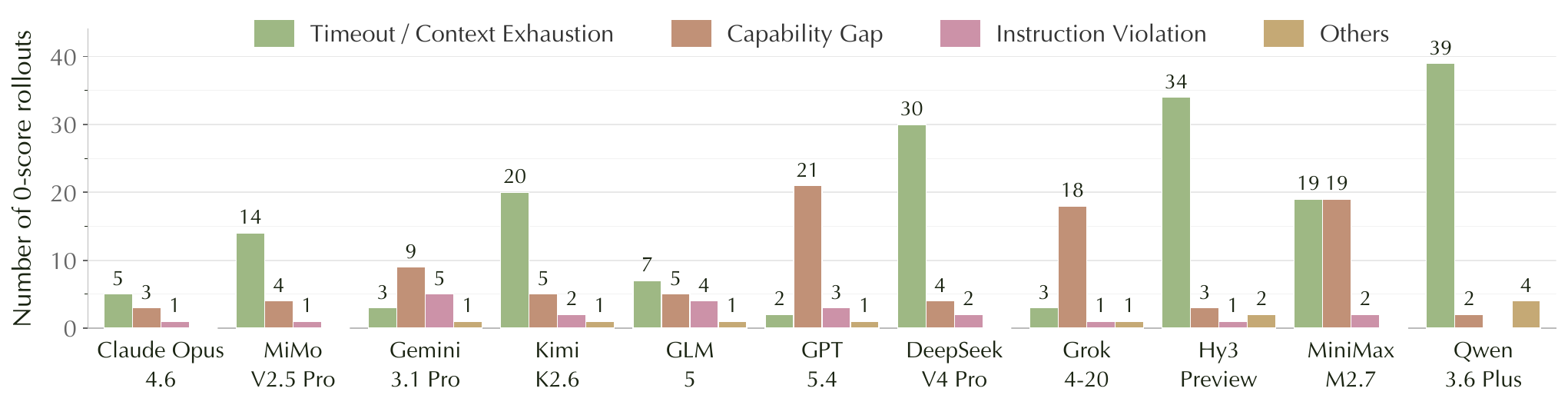}
  \caption{Distribution of zero-score rollouts by failure mode across models. For each model we manually categorized all rollouts that received a score of $0$ into four mutually exclusive failure modes.}
  \label{fig:failure-modes}
\end{figure}

\subsection{Failure Case Analysis}
\label{sec:failure-analysis}

Despite the progress of models on \bench tasks, a substantial fraction of rollouts still receive a score of zero. Understanding the underlying causes of these failures is critical for identifying the current limitations of models and agents on ultra long-horizon tasks. To move beyond aggregate performance metrics and analyze \emph{why} agents fail, we manually inspected all $302$ zero-score rollouts across the $11$ evaluated models and grouped them into four mutually exclusive failure modes. These categories together account for the entire set and are defined as follows:

\begin{itemize}[topsep=0pt, itemsep=2pt, leftmargin=*]
  \item \textbf{Timeout / Context Exhaustion.} The agent never produces a final submission within the time budget. This includes both the standard \texttt{AgentTimeoutError} from the harness and individual LLM calls that hang for $1{,}500+$ seconds due to long reasoning.

  \item \textbf{Capability Gap.} The agent submits a solution, but the verifier gives it a score of $0$. This covers incorrect outputs, sub-threshold scores, early give-ups with no improvement over the baseline, or missing required files (e.g., no LoRA adapter provided in LoRA training tasks).

  \item \textbf{Instruction Violation.} The submitted solution breaks explicit task constraints (e.g., using banned APIs like \texttt{cudaMemcpy} on \texttt{ntt\_butterfly\_cuda}, importing disallowed modules, modifying protected reference files, or leaving extra files in the workspace). The verifier scores these as $0$ regardless of correctness.

  \item \textbf{Others.} Upstream issues unrelated to the agent, such as internal server errors, malformed responses, or sandbox crashes caused by illegal operations.
\end{itemize}

Figure~\ref{fig:failure-modes} shows the distribution of these failure modes per model. In what follows, we summarize two key findings.

\textbf{Models struggle to calibrate exploration with the remaining time budget.} Models exhibit a pronounced lack of time awareness, falling into two distinct behavioral patterns: some terminate far too early, while others continue iterating until the budget is exhausted without ever submitting a final solution. A timeout-dominated group, including \texttt{deepseek-v4-pro}, \texttt{hunyuan-3-preview}, and \texttt{qwen-3.6-plus}, frequently fails to reach submission, instead consuming the entire budget through excessive iteration and repeated re-prompting. At the opposite extreme, \texttt{gpt-5.4} and \texttt{grok-4-20} often submit after only minimal exploration, resulting in consistently low scores despite substantial remaining budget. 

In addition, we observe a failure mode that is exclusive to the open-weight models in our roster: the agent enters an extremely long reasoning chain and exhausts the two-hour budget after emitting only a handful of actions. This shows up for \texttt{kimi-k2.6} on \texttt{toy\_isa\_opt} (all three trials time out at 2--11 agent steps, scoring 0) and most starkly for \texttt{deepseek-v4-pro} on the CUDA subset (9 of 12 trials submit fewer than 10 actions before the agent timeout). By contrast, none of the closed-weight models exhibits this pattern.

\textbf{Instruction violations persist even among strong closed-source models.}
Although they represent only a modest fraction of all failures, instruction violations are heavily concentrated in \texttt{gemini-3.1-pro} ($5$ cases) and \texttt{glm-5} ($4$ cases). Notably, a single task (\texttt{ntt\_butterfly\_cuda}) accounts for half of all violations. These findings, along with model-specific patterns observed in \texttt{gemini-3.1-pro} and \texttt{grok-4-20}, suggest that robust instruction following remains a significant challenge even for capable frontier models.

\subsection{Harness Ablation Analysis}
\label{app:harness}

The choice of agent harness is frequently treated as a mere implementation detail when reporting model capabilities. To examine the validity of this assumption, we re-evaluated four models (\texttt{mimo-v2.5}, \texttt{gpt-5.4}, \texttt{deepseek-v4-flash}, and \texttt{kimi-k2.6}) on two alternative harnesses: \texttt{mini-swe-agent}~\citep{yang2024sweagent} and \texttt{pi-mono}~\citep{pi_mono}. We used 25 CPU tasks from \texttt{system\_optimization} and \texttt{puzzle\_and\_challenge} and compared the results to our default \texttt{terminus-2} Harness. Since the original \texttt{mini-swe-agent} was designed primarily for patch-based editing rather than sustained iterative optimization, models tended to submit prematurely. We therefore augmented it with a custom system prompt (shown below) that explicitly encourages aggressive, persistent performance engineering and discourages early termination.

\vspace{0.5em}
\begin{promptbox}{\faIcon{terminal}\ \texttt{modified mini-swe-agent} system prompt}
\ttfamily\footnotesize\raggedright
You are an aggressive performance-engineering agent operating in a Linux container.\par\smallskip
The codebase under \texttt{/app} is already correct. Your job is to minimise a metric described in the task instruction. Higher quality optimisation = higher reward.\par\smallskip
\textbf{\textcolor{narragreendeep}{Workflow:}}
\begin{enumerate}\itemsep1pt\parskip0pt
\item Read the task instruction carefully. Note the metric, the baseline score, the reference (target) score, and any rules.
\item Read the existing \texttt{solve.*} and \texttt{main.*} files to understand the harness.
\item Run the existing build/test once to confirm baseline.
\item \textbf{Iteratively optimise.} After each change: build, run the local verifier, check the metric. If correctness fails, fix it. If the metric improved, keep going---try further optimisations. If it regressed, revert and try something else.
\item Aim for the reference score (or better). \textbf{DO NOT submit at first pass}---keep optimising until you have spent a meaningful fraction of your budget OR you reach the reference.
\item Only submit your solution once you have exhausted
reasonable optimisation ideas AND verified the final solution still passes correctness.
\end{enumerate}
\end{promptbox}
\vspace{0.5em}

\begin{figure}[htbp]
\centering
\includegraphics[width=\textwidth]{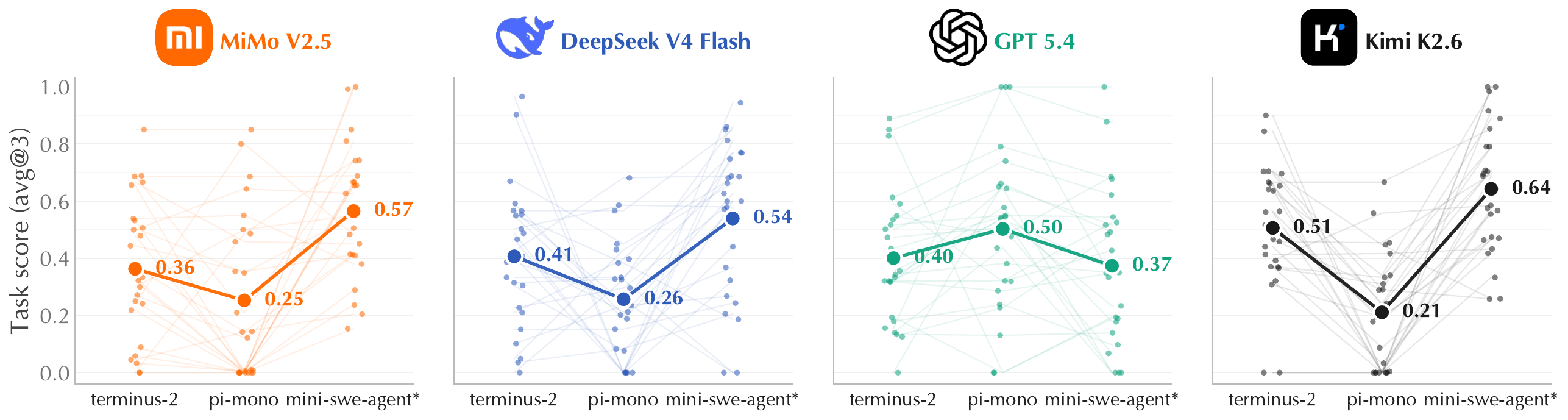}
\caption{Per-task scores across three harnesses (\texttt{terminus-2}, \texttt{pi-mono}, \texttt{mini-swe-agent*}; $^{*}$ denotes our custom optimisation-oriented system prompt), with one panel per model. Thin colored lines trace a single task's Avg@3 across harnesses; bold lines and large markers show per-harness means (also labeled).}
\label{fig:harness_slope}
\end{figure}

Figure~\ref{fig:harness_slope} reveals that harness-induced variance can be comparable to model-induced variance. Mean scores for the same model shift by as much as $\Delta=0.43$ across harnesses (e.g., \texttt{kimi-k2.6}: $0.21\to 0.64$ from \texttt{pi-mono} to \texttt{mini-swe-agent*}). Relative rankings are non-transitive: \texttt{pi-mono} favors \texttt{gpt-5.4}, \texttt{mini-swe-agent*} favors \texttt{kimi-k2.6}, and \texttt{terminus-2} lies roughly in between. Moreover, even within a fixed (model, harness) pair, per-task scores exhibit large spread, indicating substantial task-level reordering across harnesses. To ensure a fair evaluation, we selected \texttt{terminus-2}, one of the most widely adopted and well-established agent harnesses in current coding agent evaluations~\cite{jimenez2024swebench}, as the default harness for \bench{}. This provides a strong, standardized baseline for long-horizon agent evaluation.

\begin{figure}[t]
\centering
\includegraphics[width=\textwidth]{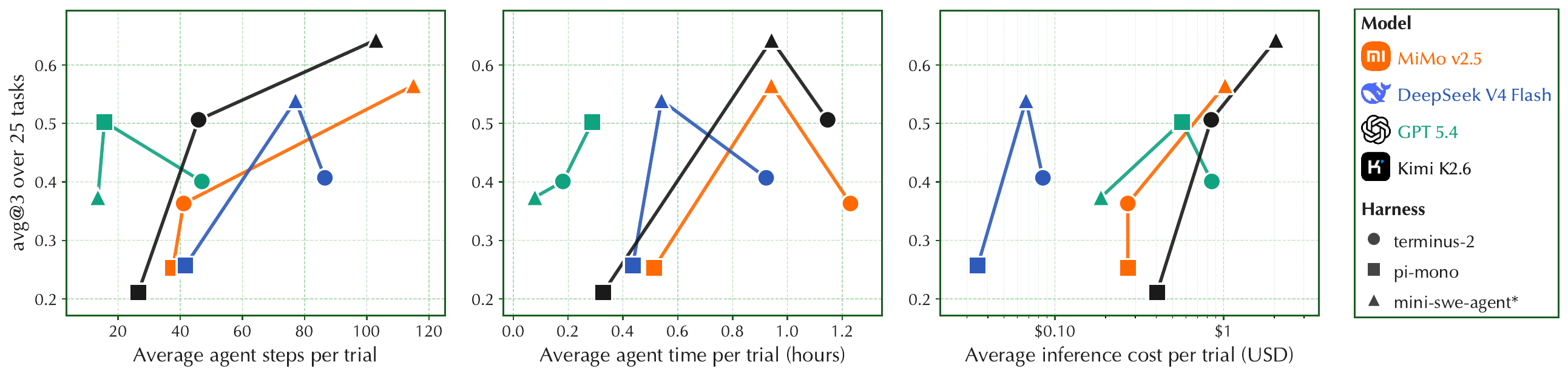}
\caption{Score versus compute spend across three harnesses on the 25-task subset. Each polyline connects the three harness points for a single model. Marker shapes denote harnesses ($\bullet$~\texttt{terminus-2}, $\blacksquare$~\texttt{pi-mono}, $\blacktriangle$~\texttt{mini-swe-agent*}). Cost is shown on a log scale. Higher and further left is Pareto-better.}
\label{fig:harness_cost}
\end{figure}

To further disentangle score differences from compute usage, Figure~\ref{fig:harness_cost} examines the score--compute trade-off across the same 25-task subset.
Three key patterns emerge: 
\textbf{(i) Harness choice is also a spending choice.} Per-trial inference cost varies by more than $5\times$ across harnesses for the same model (e.g., \texttt{kimi-k2.6}: \$0.40 under \texttt{pi-mono} vs.\ \$2.05 under \texttt{mini-swe-agent*}), primarily because different harnesses encourage vastly different iteration efforts before termination.
\textbf{(ii) Score--cost rankings diverge from score-only rankings.} Smaller or less capable base models paired with persistent harnesses (e.g., \texttt{deepseek-v4-flash} on \texttt{mini-swe-agent*}, achieving $0.54$ at $\sim$\$0.07/trial) can dominate more capable models on the cost-adjusted Pareto frontier, even though they score lower in isolation.
\textbf{(iii) Harness design amplifies or dampens specific model strengths.} Under the iterative \texttt{mini-swe-agent*}, the less capable models \texttt{deepseek-v4-flash} and \texttt{mimo-v2.5} benefit the most ($+0.13$ and $+0.20$ relative to \texttt{terminus-2}), while \texttt{gpt-5.4} slightly declines ($-0.03$). In contrast, under the lightweight \texttt{pi-mono}, the pattern reverses: only \texttt{gpt-5.4} maintains solid performance ($0.50$, the highest in that harness), while the other three models collapse to $0.21$--$0.27$. In short, the iterative harness (i.e., \texttt{mini-swe-agent*}) let the agent recover through trial-and-error what a strong single-shot reasoner would solve in one pass; models weaker at one-shot reasoning therefore gain the most, while a model that excels at one-shot reasoning gains little or loses ground.

Taken together, these findings suggest that \textbf{harness design itself is a promising direction for future research}: carefully tuned harnesses, by offering more iteration headroom for smaller models and tighter, high-quality patch loops for stronger instruction-followers, could close a substantial portion of the performance gap between weak and strong base models without any changes to the underlying models.

\subsection{More Analysis}
\label{sec:more-analysis}

In Appendix~\ref{app:more_analysis}, we present two more complementary analyses that provide additional insight into agent performance on \bench{}. First, we analyze generational improvements within model families while holding the harness fixed, and observe modest gains in most cases, although these improvements are uneven across sub-domains. Second, we study across-trial stability and find that capability and reliability are correlated but distinct dimensions. \texttt{claude-opus-4.6} performs strongly on both axes, whereas many lower-performing models also exhibit substantial variance, making single-trial evaluation unreliable and artificially inflating the perceived strength of noisy models when only Best@3 is considered.

\section{Related Work}
\label{sec:related}

\paragraph{Static and short-horizon agent benchmarks.}
The majority of public frontier-model evaluations are still single-turn or terminal-state. Coding suites such as HumanEval~\citep{chen2021codex}, LiveCodeBench~\citep{jain2025livecodebench}, BigCodeBench~\citep{zhuo2025bigcodebench}, and SWE-bench~\citep{jimenez2024swebench} score one-shot generation or one-edit-one-submission, while saturation-resistant variants like MMLU-Pro~\citep{wang2024mmlupro} and LiveBench~\citep{white2024livebench} raise difficulty without changing the framing. Multi-step agent benchmarks such as GAIA~\citep{mialon2023gaia}, AgentBench~\citep{liu2023agentbench}, and Terminal-Bench~\citep{merrill2026terminalbench} extend trajectories to several tool calls, but still grade only a final artifact or terminal state. None of them elevate sustained empirical iteration under a quantitative metric to the primary object of evaluation.

\paragraph{Long-horizon optimization and research-agent benchmarks.}
A growing family of benchmarks studies agents on realistic multi-hour ML and engineering workflows. MLE-Bench~\citep{chan2024mlebench}, RE-Bench~\citep{wijk2024rebench}, PaperBench~\citep{starace2025paperbench}, PostTrainBench~\citep{rank2026posttrainbench}, and AIRS-Bench~\citep{lupidi2026airsbench} target ML research pipelines, while KernelBench~\citep{ouyang2025kernelbench}, FrontierCS~\citep{mang2025frontiercs}, and Frontier-Eng~\citep{chi2026frontier} probe systems, kernel, and real-world engineering optimization. These works represent important progress toward agentic research, but each focuses on a narrow domain and typically grades only the final score or final patch, leaving the trajectory itself unmeasured. \bench{} departs from this convention by spanning four heterogeneous categories (system optimization, puzzles, model development, and CUDA kernels) under a single calibrated, hack-resistant scoring protocol.

\paragraph{Closed-loop agent frameworks and training environments.}
Numerous frameworks have been proposed for closed-loop software engineering (SWE-agent~\citep{yang2024sweagent}, OpenHands~\citep{wang2024openhands}, Aider~\citep{gauthier2024aider}) and open-ended scientific iteration (The AI Scientist~\citep{lu2024aiScientist}). Gym-style environments such as SWE-Gym~\citep{pan2024swegym}, R2E-Gym~\citep{jain2025r2egym}, and MLGym~\citep{nathani2025mlgym} further enable iterative training and evaluation on engineering and ML tasks. The most striking demonstrations of sustained empirical optimization, including AlphaEvolve~\citep{novikov2025alphaevolve} and \citeauthor{karpathy2026autoresearch}'s AutoResearch agent~\citep{karpathy2026autoresearch}, are tightly coupled to bespoke harnesses, tools, and search strategies, which makes the underlying model's contribution difficult to isolate from system-level engineering. \bench{} instead fixes the harness (\texttt{terminus-2}), action interface, task definitions, budget rules, and scoring function across all evaluated models, enabling direct, apples-to-apples comparison of long-horizon optimization capability under identical conditions.

\section{Conclusion}

We introduced \bench{}, a benchmark for evaluating frontier models on ultra long-horizon research and engineering tasks that require sustained iteration over hours rather than minutes. By enforcing ultra long-horizon tasks, continuous calibrated scoring, and strong anti-hacking safeguards, \bench{} reveals that raw capability alone is insufficient for these tasks: the dominant predictor of success is an agent's willingness to persistently evaluate, edit, and iterate over extended horizons. \texttt{claude-opus-4.6} demonstrates this convincingly, achieving a commanding lead through long, steady optimization trajectories while most other frontier models, including several proprietary models, terminate prematurely or exhaust their budgets without submitting. These results highlight the critical need for future models and agents to prioritize time awareness, persistence, and more effective harness design. We release the full benchmark, evaluation harness, and task artifacts to accelerate progress toward truly capable ultra long-horizon agents.
\section*{Limitations and Broader Impact}\label{sec:limitations}

\bench focuses on executable system and machine learning engineering workflows, and should therefore be understood as a benchmark for measurable auto-research rather than for scientific discovery in its broadest sense. Because long-horizon evaluation inherently depends on multi-hour execution, API interactions, GPU workloads, and the surrounding execution stack, \bench reports trajectory analysis, resource consumption, and final performance jointly rather than treating benchmark score as a standalone metric. We hope this protocol supports more diagnostic, cost-aware, and reproducible evaluation of auto-research agents as the field moves from static answers toward iterative empirical work.

% Acknowledgments 
\section*{Acknowledgments}
The authors thank Professor Erik Brynjolfsson (Stanford University) for valuable discussions that helped sharpen the framing of this work, and colleagues across our affiliated institutions for constructive feedback on earlier drafts.

% --- Bibliography ---
\bibliography{refs}

% ============================================================================
%  Appendix
% ============================================================================
\clearpage
\appendix
\section{Task Specifications}
\label{app:tasks}

\subsection{Task Descriptions}
\label{app:task-descriptions}

This subsection lists all 36 tasks in \bench{}, grouped by category. Each entry gives the implementation language, a difficulty tier (1 = textbook-classic optimization with well-known techniques; 2 = bespoke, domain-specific, or research-style), and a short description.

\paragraph{System Optimization (15 tasks).}\leavevmode\par\medskip

\begin{taskcard}{csystem}{aes128\_ctr}{C, tier 1}
Optimizes AES-128 CTR mode encryption of 256\,MiB in single-threaded C using hardware acceleration and SIMD intrinsics. The challenge is maximizing throughput by leveraging AES-NI instructions and pipelining to minimize latency-bound operations.
\end{taskcard}

\begin{taskcard}{csystem}{agent\_tool\_routing}{Python, tier 2}
Optimizes a Python lexical router that returns the top-10 tool schemas for natural-language agent queries while satisfying $\mathrm{MRR}@10 \geq 0.82$ and $\mathrm{Recall}@10 \geq 0.94$. The challenge is replacing a full-scan baseline with a weighted inverted index using IDF and stopword filtering, all in single-threaded stdlib Python with no external packages.
\end{taskcard}

\begin{taskcard}{csystem}{bm25\_search\_go}{Go, tier 2}
Optimizes BM25 search engine queries over a synthetic corpus using goroutines and stdlib. The challenge is implementing efficient inverted indexes and ranking algorithms to minimize query latency while managing concurrent goroutine overhead.
\end{taskcard}

\begin{taskcard}{csystem}{bvh\_raytracer}{C++, tier 1}
Builds a BVH for a 4\,096-triangle scene and optimizes ray-triangle intersection for $638{\times}638$ primary rays. The challenge is constructing a cache-friendly BVH hierarchy and using SIMD vectorization to batch ray intersection tests while minimizing traversal overhead.
\end{taskcard}

\begin{taskcard}{csystem}{concurrent\_kv\_wal}{Go, tier 2}
Optimizes a write-ahead log (WAL) backed key-value store handling mixed read/write/query workloads from 4 concurrent goroutines. The challenge is reducing lock contention and I/O bottlenecks while maintaining consistency and durability guarantees.
\end{taskcard}

\begin{taskcard}{csystem}{fft\_rust}{Rust, tier 2}
Optimizes the Fast Fourier Transform of a 32\,768-length real signal in pure Rust. The challenge is implementing cache-efficient FFT algorithms with proper memory layout and minimizing complex number arithmetic operations.
\end{taskcard}

\begin{taskcard}{csystem}{flash\_attention}{C, tier 1}
Optimizes scaled dot-product attention ($n=4096$, $d=64$) while avoiding full $n{\times}n$ matrix allocation through tiled computation. The challenge is implementing memory-efficient attention with SIMD softmax to minimize bandwidth and maintain numerical stability.
\end{taskcard}

\begin{taskcard}{csystem}{gaussian\_blur}{C, tier 1}
Applies a $17{\times}17$ Gaussian blur to a $4096{\times}4096$ image 5 times sequentially in single-threaded C. The challenge is exploiting separable filtering and SIMD vectorization while maintaining optimal cache locality across multiple passes.
\end{taskcard}

\begin{taskcard}{csystem}{hash\_join}{C, tier 1}
Executes an inner equi-join of 20K and 5M row tables on an integer key using hash-join in single-threaded C. The challenge is building cache-friendly hash tables with minimal collisions and using SIMD prefetching to hide memory latency.
\end{taskcard}

\begin{taskcard}{csystem}{levenshtein\_distance}{C, tier 1}
Optimizes a C \texttt{levenshtein()} routine to compute exact edit distances over 1\,000\,000 deterministic lowercase ASCII string pairs of lengths 16--64. The challenge is replacing the $O(l_a \cdot l_b)$ two-row DP with a Myers bit-parallel algorithm that packs each row into a \texttt{uint64\_t} and performs $\sim$8 branchless 64-bit ops per column, while staying single-threaded and bit-exact.
\end{taskcard}

\begin{taskcard}{csystem}{radix\_sort}{C, tier 1}
Sorts 50M random uint32 values as fast as possible using radix sort in single-threaded C. The challenge is maximizing memory bandwidth efficiency through careful radix digit selection, buffering strategies, and minimizing cache misses.
\end{taskcard}

\begin{taskcard}{csystem}{regex\_engine}{Rust, tier 2}
Compiles regex patterns and searches 100K haystacks using pure Rust without external crates. The challenge is building efficient NFA/DFA structures with backtracking optimization to minimize pattern matching latency across large text volumes.
\end{taskcard}

\begin{taskcard}{csystem}{sha256\_throughput}{C, tier 1}
Hashes a 512\,MiB buffer with SHA-256 as fast as possible in single-threaded C using intrinsics. The challenge is maximizing throughput by exploiting parallelism within SHA-256 compression rounds and maintaining optimal instruction pipelining.
\end{taskcard}

\begin{taskcard}{csystem}{sstable\_compaction\_rs}{Rust, tier 2}
Optimizes LSM-style SSTable compaction pipeline merging sorted tables in pure Rust. The challenge is implementing efficient multi-way merge with prefix compression and block boundary optimization to minimize I/O and CPU overhead.
\end{taskcard}

\begin{taskcard}{csystem}{z\_order\_range\_scan}{Rust, tier 2}
Optimizes a Rust 2D rectangular range-count index over a 16-bit coordinate multiset answering many \texttt{count\_in(xlo, ylo, xhi, yhi)} queries per case. The challenge is replacing the linear per-query scan with a Z-order/Morton spatial key sort plus a skip-step forward scan that leaps over coordinate runs outside the query rectangle, all in safe stdlib Rust with no threading or unsafe.
\end{taskcard}

\paragraph{Puzzle and Challenge (10 tasks).}\leavevmode\par\medskip

\begin{taskcard}{cpuzzle}{adaptive\_compression}{Python, tier 2}
Implements a byte-level Predictor in \texttt{/app/predictor.py} whose predict/update loop minimizes average bits-per-byte across 9 hidden families of sequences (Markov chains, periodic, bracketed, recurrences, regime switches, RLE, random), with a 5.0 bpb baseline (order-1 Markov) and a 3.8 bpb reference (PPM-style blended context mixer with match model and period detector). The challenge is that no single model wins on every family, so the agent must build an adaptive context-mixing compressor that tracks higher-order statistics, detects periodicity, and switches regimes online without ever emitting an invalid distribution.
\end{taskcard}

\begin{taskcard}{cpuzzle}{adversarial\_splay}{Python, tier 2}
Writes 4\,096 key accesses (each in $[0,1023]$) to \texttt{/app/accesses.txt} that maximize the rotation count of a deterministic top-down Sleator--Tarjan splay tree built by inserting $0..1023$ in order, against a uniform-random baseline of 48\,656 rotations and a bit-reversal-cycle reference of 67\,008 rotations. The challenge is to design a sequence that simultaneously defeats the Sequential, Working-Set, and Dynamic-Finger access bounds, which requires reasoning about how zig / zig-zig / zig-zag steps reshape the tree across the full 4\,096-step amortized trajectory.
\end{taskcard}

\begin{taskcard}{cpuzzle}{discover\_sorting}{Python, tier 1}
Implements \texttt{generate\_network(16)} in \texttt{/app/solve.py} to emit a correct 16-input sorting network with as few comparators as possible, against a Batcher bitonic baseline of 80 comparators and a known-optimal reference of 60 (Dobbelaere 2019). The challenge is that constructing networks below the bitonic count requires either reproducing the literature's hand-tuned optimal layouts or running a careful combinatorial search verified by the 0--1 principle over all 65\,536 binary inputs.
\end{taskcard}

\begin{taskcard}{cpuzzle}{fredkin\_sort\_network}{custom, tier 2}
Rewrites \texttt{/app/circuit.txt} to implement a 4-input 2-bit-wide stable reversible sorting network using only \texttt{x}/\texttt{cx}/\texttt{ccx}/\texttt{fred} gates, restoring all scratch wires to zero, against a 128-gate baseline and an 88-gate streamlined reference. The challenge is that reversible computation forbids destructive overwrites, so every comparator must be built from carefully shared intermediate terms and then fully uncomputed, making gate reduction a tight routing-and-cleanup puzzle.
\end{taskcard}

\begin{taskcard}{cpuzzle}{resnet\_bit\_flip}{Python, tier 2}
Implements \texttt{find\_bit\_flips()} in \texttt{/app/solve.py} to return the smallest set of float32 bit edits to a frozen $\sim$77K-parameter MiniResNet that drives CIFAR-10 top-1 accuracy below 12\% (baseline 95 flips via top-$K$ sign-bit flipping, reference 40 flips via 1P-DNL $|w|{\cdot}|\nabla|$ saliency from Guzm\'an et al.\ 2025). The challenge is that the $|w|{\le}10$ magnitude cap and finite-output requirement rule out brute-force exponent attacks, so the agent must combine gradient/saliency analysis with careful search to identify the few weight bits whose flip cascades through the network into class collapse.
\end{taskcard}

\begin{taskcard}{cpuzzle}{safety\_router}{Python, tier 1}
Optimizes \texttt{/app/train.py} to train a 2-layer MLP refusal router with as few parameters as possible while passing private-split gates of accuracy $\geq 0.64$, unsafe recall $\geq 0.66$, and safe recall $\geq 0.57$, against a 16\,641-parameter baseline (128-hidden) and a 2\,081-parameter reference. The challenge is shrinking the hidden width and tuning thresholds aggressively without violating the asymmetric refusal-vs-answer recall gates, since under-compressing wastes parameters but over-compressing collapses one of the class recalls below threshold and zeroes the reward.
\end{taskcard}

\begin{taskcard}{cpuzzle}{smallest\_game\_player}{Python, tier 2}
Implements \texttt{train()}/\texttt{predict()} in \texttt{/app/solve.py} to learn optimal moves for $4{\times}4$ gravity Connect-3 with as few learnable parameters as possible while keeping hidden-test accuracy $\geq 95\%$, against a 17\,924-parameter tiny-transformer baseline and a 913-parameter weight-shared per-column scoring reference. The challenge is that explicit minimax/search and hardcoded lookup tables are forbidden, so the agent must hand-design tactical features and a weight-shared scoring head compact enough to encode the game's tactics in under a thousand parameters.
\end{taskcard}

\begin{taskcard}{cpuzzle}{stack\_machine\_golf}{custom, tier 2}
Rewrites \texttt{/app/program.stk} to compute a 256-element integer dot product on a register-less stack VM in as few executed instructions as possible, against a 5\,132-instruction memory-scratched-loop baseline and a 3\,530-instruction $4{\times}$-unrolled stack-resident-accumulator reference. The challenge is that the pure stack ISA has no registers, so every loop-counter and pointer manipulation costs extra \texttt{dup}/\texttt{swap}/\texttt{over} instructions, and instruction-count gains require carefully unrolling and keeping the running accumulator on the stack across the loop body.
\end{taskcard}

\begin{taskcard}{cpuzzle}{toy\_isa\_opt}{custom, tier 2}
Rewrites \texttt{/app/program.s} to compute a 512-element integer dot product on the PINC scoreboard pipeline in as few simulated cycles as possible, against a 9\,220-cycle naive single-accumulator baseline and a 2\,954-cycle reference using $4{\times}$ unrolling with 4 independent accumulators and interleaved loads. The challenge is that the in-order pipeline stalls on RAW hazards through 5-cycle loads and 5-cycle muls, so cycle reduction requires software pipelining, multi-accumulator unrolling, and careful instruction interleaving to fully hide latency.
\end{taskcard}

\begin{taskcard}{cpuzzle}{vliw\_scheduler}{custom, tier 1}
Implements \texttt{vliw\_schedule()} in \texttt{/app/solve.c} to pack 3\,000 adversarially ordered ALU/MUL/MEM ops (ending in 60 chains of 10 dependent muls) into 3-slot VLIW bundles obeying $1$/$3$/$4$-cycle latencies, against a 4\,080-cycle one-op-per-bundle baseline and a 1\,300-cycle critical-path list-scheduling reference. The challenge is that the long MUL chains form the critical path but appear last in input order, so a competitive scheduler must build the full dependence DAG, compute latency-weighted heights, and use priority-driven list scheduling to fill all three slots per cycle.
\end{taskcard}

\paragraph{Model Development (7 tasks).}\leavevmode\par\medskip

\begin{taskcard}{cmodel}{data\_select\_ifeval}{Python, tier 2}
Selects up to 5\,000 training samples from a 50\,000-sample pool spanning 19 sources to maximize IFEval prompt-strict accuracy after LoRA fine-tuning Qwen2.5-3B-Instruct on a single H100 within 8 hours. The challenge is that the base model is already instruction-tuned so most pool data is unhelpful or harmful, and the agent must reason about which sources (persona\_if, coconot, wildchat, math, code, multilingual, safety) actually transfer to format-constraint following without being able to inspect IFEval directly.
\end{taskcard}

\begin{taskcard}{cmodel}{flux2\_klein\_lora}{Python, tier 1}
Trains a LoRA adapter on the FLUX.2 klein 9B DiT with musubi-tuner over 15 concept images on a single L40S (48\,GB) within 4 hours, optimizing a composite of CLIP image similarity, DINO structural similarity, and CLIP text alignment across 8 eval prompts. The challenge is that the shipped config OOMs and the agent must jointly fix the OOM (gradient checkpointing, blocks-to-swap, batch size, resolution), correct the model version, and tune LoRA rank, learning rate, and timestep sampling for FLUX.2's flow-matching schedule.
\end{taskcard}

\begin{taskcard}{cmodel}{grpo\_multisource}{Python, tier 1}
Fine-tunes Qwen2.5-VL-7B (4-bit) with GRPO over Geometry3K, MathVision, and ChartQA to maximize MathVista accuracy on a single L40S in 8 hours, subject to a retention gate that zeros the score if general VQA drops more than 10\%. The challenge is mixing three heterogeneous visual-math sources, designing reward functions that are robust to formatting variance, and tuning GRPO hyperparameters (lr, num\_generations, LoRA rank, KL) so the policy generalizes to MathVista without forgetting base capabilities.
\end{taskcard}

\begin{taskcard}{cmodel}{llm\_online\_serving}{Python, tier 2}
Optimizes the SimpleLLM online serving engine for gpt-oss-20b (21B MoE, 3.6B active) on a single H100 to maximize a $50/50$ blend of throughput and inverse mean completion time over 96 Poisson-arriving requests within 2 hours. The challenge is editing only the engine layer (continuous batching, slot management, async request queue) while kernels and the model are frozen, which forces gains from CUDA graph capture across batch sizes, prefill chunk budgeting, and scheduler tuning while still passing a correctness test.
\end{taskcard}

\begin{taskcard}{cmodel}{moving\_mnist\_world\_model}{Python, tier 1}
Trains a video next-frame prediction model from scratch on 8\,000 Moving MNIST clips (20 frames, $64{\times}64$) to maximize PSNR over autoregressive 10-step rollouts on a held-out test split, using a single H100 within 4 hours. The challenge is selecting an architecture and training schedule that produces sharp multi-step rollouts without ground-truth conditioning, since naive ConvLSTMs blur quickly under autoregressive error accumulation and the test split is unseen at design time.
\end{taskcard}

\begin{taskcard}{cmodel}{multilingual\_ocr}{Python, tier 1}
Fine-tunes DeepSeek-OCR (3B) with Unsloth LoRA on 9\,000 mixed Persian and Bengali synthetic-print images to minimize average character error rate across 400 held-out images on a single L40S within 8 hours. The challenge is balancing two scripts with very different glyph shapes under a single LoRA, choosing rank, learning rate, and step count to escape a weak $r{=}16$ / 60-step baseline without overfitting on the synthetic distribution.
\end{taskcard}

\begin{taskcard}{cmodel}{scaling\_law}{Python, tier 1}
Pretrains a transformer language model from scratch on WikiText-103 ($\sim$118M tokens, GPT-2 tokenizer) with LitGPT to minimize test perplexity on a single H100 within 12 hours, with eval fixed at \texttt{seq\_length=1024}. The challenge is choosing a compute-optimal point in the (model size, tokens, steps) space, modernizing the architecture (RMSNorm, SwiGLU, GQA, RoPE), and configuring bf16 plus \texttt{torch.compile}, cosine LR with warmup, and batch settings to close the perplexity gap from a $\sim$95 pythia-14m baseline to a $\sim$23 reference at $\sim$179M Llama-style parameters.
\end{taskcard}

\paragraph{CUDA (4 tasks).}\leavevmode\par\medskip

\begin{taskcard}{ccuda}{huffman\_canonical\_decode\_cuda}{CUDA, tier 2}
Decodes 2\,048 independent canonical-Huffman bitstreams (each 65\,536 bytes of payload, max 16-bit codes) on a single H100 by writing a custom kernel with shared-memory primary/fallback LUTs, warp cooperation, and rolling bit registers. The challenge is that canonical Huffman decoding is inherently serial within a stream (each symbol's length depends on the previous), so reaching the reference ($\sim$3\,ms vs.\ 56\,ms baseline) requires multi-level shared-memory LUTs, warp-cooperative output packing, and unrolled multi-symbol-per-iteration loops without using Thrust, CUB, or any compression library.
\end{taskcard}

\begin{taskcard}{ccuda}{icp\_correspondence\_step\_cuda}{CUDA, tier 2}
Runs one Iterative-Closest-Point correspondence step on a single H100 over a source cloud of $N{=}200\,000$ and target cloud of $M{=}500\,000$ fp32 points, returning the $3{\times}3$ cross-covariance $H$, squared-error sum, and valid-pair count using a prebuilt KD-tree with $d_\mathrm{max}=0.05$ distance gating. The challenge is that brute-force $O(NM)$ NN is far too slow ($\sim$65\,ms baseline), so reaching the $\sim$0.28\,ms reference requires register-stack KD-tree DFS with AABB pruning and ordered traversal, on-GPU centroid computation, and warp-cooperative fp64 covariance reduction with \texttt{\_\_shfl\_down\_sync}, all without Thrust/CUB/cuBLAS or any spatial-index library.
\end{taskcard}

\begin{taskcard}{ccuda}{msm\_pippenger\_bls12\_381\_cuda}{CUDA, tier 2}
Computes a BLS12-381 G1 multi-scalar multiplication $Q = \sum s_i P_i$ over $N{=}262\,144$ affine points (Montgomery-form 384-bit coordinates) and 256-bit scalars on a single H100, the dominant kernel in zkSNARK provers like Groth16 and PLONK. The challenge is that naive double-and-add is $\sim$9$\times$ slower than reference (532\,ms vs.\ 59\,ms), so reaching the target requires implementing the Pippenger bucket method ($c{=}8$, 32 windows $\times$ 255 buckets) with chunked accumulate / tree-reduce / descending running-sum window combine and CIOS Montgomery field arithmetic in inline PTX, with no third-party finite-field, ECC, or zkSNARK library and no in-kernel allocations.
\end{taskcard}

\begin{taskcard}{ccuda}{ntt\_butterfly\_cuda}{CUDA, tier 1}
Applies an in-place batched forward Number Theoretic Transform over the Goldilocks prime field ($p = 2^{64} - 2^{32} + 1$) on a single H100 for batch=256 rows of $n{=}65\,536$ uint64 elements, producing natural-order output bit-exactly matching the reference. The challenge is that a textbook iterative Cooley--Tukey kernel with per-thread twiddle re-derivation and integer modulo runs $\sim$85$\times$ slower than reference (110\,ms vs.\ 1.28\,ms), so reaching the target requires a precomputed twiddle table, shared-memory tiled butterflies for the early stages with global-memory butterflies for the rest, and Goldilocks-specific 128-to-64-bit modular reduction without integer division, all without cuFFT, Thrust, CUB, or any FFT library.
\end{taskcard}

\subsection{Per-Task Scoring Anchors and Gates}
\label{app:scoring}

The two scoring schemes, \emph{anchored linear} and \emph{log-stretch}, are formally defined in Section~\ref{subsec:task_formulation}. Both schemes are clipped to the interval \([0,1]\) and saturate to \(0\) whenever the agent fails the correctness check. In this section we provide, for every task, the specific metric, its direction (\(\downarrow\) for lower-is-better, \(\uparrow\) for higher-is-better), the baseline anchor \(m_\mathcal{B}\), the reference anchor \(m_\mathcal{R}\), and any task-specific feasibility gate that must be satisfied before a positive score is awarded.

Two parameter-count tasks in the Puzzle \& Challenge category (\texttt{smallest\_game\_player} and \texttt{safety\_router}) employ the degenerate linear form \(s(x)=\mathrm{clip}\big((m_\mathcal{B}-m(x))/m_\mathcal{B},0,1\big)\), which is mathematically equivalent to anchored linear scoring with an implicit reference anchor of \(m_\mathcal{R}=0\) (zero parameters). The \(m_\mathcal{R}\) values listed for these tasks in Table~\ref{tab:puzzle-rewards} therefore represent the documented strong solution rather than the scoring anchor itself. The \texttt{resnet\_bit\_flip} task additionally imposes a feasibility gate requiring the corrupted model's accuracy to fall below \(12\%\) before any positive reward is granted. All system-optimization and CUDA tasks use log-stretch scoring on a speed-up metric with a ``must beat baseline'' gate.

\begin{table}[t]
\centering
\scriptsize
\caption{System-optimization (15 tasks) and CUDA (4 tasks), all using log-stretch scoring with a ``must beat baseline'' improvement gate. 
The values \(m_\mathcal{B}\) and \(m_\mathcal{R}\) are the \emph{empirically measured} runtimes of the baseline and reference implementations on the benchmark's standardized hardware and sandbox environments (seconds for system-optimization tasks, milliseconds for CUDA tasks).}
\label{tab:sysopt-cuda-rewards}
\resizebox{\textwidth}{!}{%
\begin{tabular}{llcrrl}
\toprule
\textbf{Task} & \textbf{Metric} & \textbf{Dir.} & \textbf{$m_\mathcal{B}$} & \textbf{$m_\mathcal{R}$} & \textbf{Category} \\
\midrule
\texttt{aes128\_ctr}            & runtime\_seconds & $\downarrow$ & 3.0     & 0.10    & system-opt \\
\texttt{agent\_tool\_routing}   & runtime\_seconds & $\downarrow$ & 3.85    & 0.40    & system-opt \\
\texttt{bm25\_search\_go}       & runtime\_seconds & $\downarrow$ & 2.1     & 0.03    & system-opt \\
\texttt{bvh\_raytracer}         & runtime\_seconds & $\downarrow$ & 3.8     & 0.030   & system-opt \\
\texttt{concurrent\_kv\_wal}    & runtime\_seconds & $\downarrow$ & 9.5     & 1.1     & system-opt \\
\texttt{fft\_rust}              & runtime\_seconds & $\downarrow$ & 10.0    & 0.001   & system-opt \\
\texttt{flash\_attention}       & runtime\_seconds & $\downarrow$ & 0.75    & 0.10    & system-opt \\
\texttt{gaussian\_blur}         & runtime\_seconds & $\downarrow$ & 12.0    & 0.25    & system-opt \\
\texttt{hash\_join}             & runtime\_seconds & $\downarrow$ & 20.0    & 0.04    & system-opt \\
\texttt{levenshtein\_distance}  & runtime\_seconds & $\downarrow$ & 2.0845  & 0.3107  & system-opt \\
\texttt{radix\_sort}            & runtime\_seconds & $\downarrow$ & 4.5     & 0.35    & system-opt \\
\texttt{regex\_engine}          & runtime\_seconds & $\downarrow$ & 1.5     & 0.37    & system-opt \\
\texttt{sha256\_throughput}     & runtime\_seconds & $\downarrow$ & 2.5     & 0.15    & system-opt \\
\texttt{sstable\_compaction\_rs} & runtime\_seconds & $\downarrow$ & 0.099  & 0.041   & system-opt \\
\texttt{z\_order\_range\_scan}  & runtime\_seconds & $\downarrow$ & 2.0     & 0.02    & system-opt \\
\midrule
\texttt{huffman\_canonical\_decode\_cuda} & runtime\_ms & $\downarrow$ & 56.0  & 3.112  & CUDA \\
\texttt{icp\_correspondence\_step\_cuda}  & runtime\_ms & $\downarrow$ & 64.65 & 0.28   & CUDA \\
\texttt{msm\_pippenger\_bls12\_381\_cuda} & runtime\_ms & $\downarrow$ & 532.0 & 58.94  & CUDA \\
\texttt{ntt\_butterfly\_cuda}             & runtime\_ms & $\downarrow$ & 109.8 & 1.28   & CUDA \\
\bottomrule
\end{tabular}%
}
\end{table}

\begin{table}[t]
\centering
\scriptsize
\caption{Puzzle-and-challenge tasks (10). All use anchored linear scoring unless otherwise noted; \(m_\mathcal{B}\) and \(m_\mathcal{R}\) denote the baseline and reference targets.}
\label{tab:puzzle-rewards}
\resizebox{\textwidth}{!}{%
\begin{tabular}{llcrrl}
\toprule
\textbf{Task} & \textbf{Metric} & \textbf{Dir.} & \textbf{$m_\mathcal{B}$} & \textbf{$m_\mathcal{R}$} & \textbf{Scoring family} \\
\midrule
\texttt{discover\_sorting} & comparator\_count & $\downarrow$ & 80 & 60 & anchored linear \\
\texttt{fredkin\_sort\_network} & gate\_count & $\downarrow$ & 128 & 88 & anchored linear \\
\texttt{stack\_machine\_golf} & instruction\_count & $\downarrow$ & 5\,132 & 3\,530 & anchored linear \\
\texttt{toy\_isa\_opt} & cycles & $\downarrow$ & 9\,220 & 2\,954 & anchored linear \\
\texttt{vliw\_scheduler} & cycles & $\downarrow$ & 4\,080 & 1\,300 & anchored linear \\
\texttt{smallest\_game\_player} & total\_params & $\downarrow$ & 17\,924 & 913 & anchored linear (implicit \(m_\mathcal{R}=0\)) \\
\texttt{safety\_router} & total\_params & $\downarrow$ & 16\,641 & 2\,081 & anchored linear (implicit \(m_\mathcal{R}=0\)) \\
\texttt{resnet\_bit\_flip} & bits\_flipped & $\downarrow$ & 81 & 1 & anchored linear (\(12\%\) accuracy gate) \\
\midrule
\texttt{adaptive\_compression} & bits\_per\_byte & $\downarrow$ & 5.0 & 3.8 & log-stretch, \(5\%\) gate \\
\texttt{adversarial\_splay} & rotations & $\uparrow$ & 48\,656 & 67\,008 & log-stretch, \(1\%\) gate \\
\bottomrule
\end{tabular}%
}
\end{table}

\begin{table}[t]
\centering
\scriptsize
\caption{Model-development tasks (7), all using anchored linear scoring. 
\(^\dagger\) For \texttt{flux2\_klein\_lora}, the baseline anchor is the empirically measured no-LoRA quality score (\(\approx 0.49\)); the task configuration lists \(0.0\) as the OOM-crash floor. 
\(^\ddagger\) For \texttt{llm\_online\_serving}, the metric is a \(50/50\) throughput/latency composite that equals \(1.0\) at the baseline by construction.}
\label{tab:modeldev-rewards}
\resizebox{\textwidth}{!}{%
\begin{tabular}{llcrrl}
\toprule
\textbf{Task} & \textbf{Metric} & \textbf{Dir.} & \textbf{$m_\mathcal{B}$} & \textbf{$m_\mathcal{R}$} & \textbf{Gate} \\
\midrule
\texttt{data\_select\_ifeval} & ifeval\_prompt\_strict & $\uparrow$ & 0.38 & 0.48 & --- \\
\texttt{grpo\_multisource} & mathvista\_accuracy & $\uparrow$ & 0.20 & 0.65 & VQA retention \(\geq 0.9\) \\
\texttt{multilingual\_ocr} & avg\_cer & $\downarrow$ & 0.72 & 0.32 & --- \\
\texttt{scaling\_law} & perplexity & $\downarrow$ & 25 & 18 & --- \\
\texttt{moving\_mnist\_world\_model} & psnr & $\uparrow$ & 14 & 20 & --- \\
\texttt{flux2\_klein\_lora} & quality\_score & $\uparrow$ & 0.49\(^\dagger\) & 0.88 & beat no-LoRA \\
\texttt{llm\_online\_serving} & serving\_composite & $\uparrow$ & 1.0\(^\ddagger\) & 1.5 & correctness test \\
\bottomrule
\end{tabular}%
}
\end{table}
\section{More on Experiments}
\label{app:more_exps}

\subsection{More on Experimental Setups}
\label{app:more_exp_setups}

Table \ref{tab:model_api_providers} summarizes the developing organizations and API providers for all models evaluated on \bench.

\begin{table}[ht]
\centering
% \scriptsize
\caption{Models evaluated in \bench, with their developing organization and API provider.}
\label{tab:model_api_providers}
\begin{tabular}{lll}
\toprule
\textbf{Organization} & \textbf{Model} & \textbf{API Provider} \\
\midrule
\multicolumn{3}{l}{\textit{Main set}} \\
\cmidrule(lr){1-3}
Alibaba          & \texttt{qwen-3.6-plus}     & TokenRouter \\
Anthropic        & \texttt{claude-opus-4.6}   & Azure AI Foundry \\
DeepSeek         & \texttt{deepseek-v4-pro}   & TokenRouter \\
Google DeepMind  & \texttt{gemini-3.1-pro}    & TokenRouter \\
MiniMax          & \texttt{minimax-m2.7}      & MiniMax \\
Moonshot AI      & \texttt{kimi-k2.6}         & Cloudflare \\
OpenAI           & \texttt{gpt-5.4}           & Azure AI Foundry \\
Tencent          & \texttt{hunyuan-3-preview} & OpenRouter \\
xAI              & \texttt{grok-4-20}         & xAI \\
Xiaomi           & \texttt{mimo-v2.5-pro}     & Xiaomi \\
Zhipu AI         & \texttt{glm-5}             & TokenRouter \\
\midrule
\multicolumn{3}{l}{\textit{Ablation set}} \\
\cmidrule(lr){1-3}
Alibaba          & \texttt{qwen-3.5-plus}     & TokenRouter \\
DeepSeek         & \texttt{deepseek-v4-flash} & TokenRouter \\
MiniMax          & \texttt{minimax-m2.5}      & MiniMax \\
Moonshot AI      & \texttt{kimi-k2.5}         & Azure AI Foundry \\
Xiaomi           & \texttt{mimo-v2-pro}       & Xiaomi \\
Xiaomi           & \texttt{mimo-v2.5}         & Xiaomi \\
\bottomrule
\end{tabular}
\end{table}

\subsection{Detailed Experimental Results}
\label{app:detailed_exp_results}

Table~\ref{tab:per_task} and \ref{tab:per_task_best} reports the per-task average score and best score across three independent trials for each of the 11 frontier models in our main set, grouped by sub-domain. Cell shading uses the same diverging green to pink scale as Table~\ref{tab:main}, centred at $0.5$, so individual task strengths and weaknesses are visible at a glance. 

\begin{table*}[h]
\centering
\footnotesize
\setlength{\tabcolsep}{1pt}
\renewcommand{\arraystretch}{1.15}
\setlength{\aboverulesep}{1pt}
\setlength{\belowrulesep}{1pt}
\resizebox{\textwidth}{!}{%
\begin{tabular}{@{}p{0.22\textwidth}>{\centering\arraybackslash}p{0.0673\textwidth}>{\centering\arraybackslash}p{0.0673\textwidth}>{\centering\arraybackslash}p{0.0673\textwidth}>{\centering\arraybackslash}p{0.0673\textwidth}>{\centering\arraybackslash}p{0.0673\textwidth}>{\centering\arraybackslash}p{0.0673\textwidth}>{\centering\arraybackslash}p{0.0673\textwidth}>{\centering\arraybackslash}p{0.0673\textwidth}>{\centering\arraybackslash}p{0.0673\textwidth}>{\centering\arraybackslash}p{0.0673\textwidth}>{\centering\arraybackslash}p{0.0673\textwidth}@{}}
\toprule
\textbf{Task} & \shortstack[c]{\includegraphics[height=2.4ex]{logos/claude-opus-4.6.png}\\{\scriptsize\textbf{Claude}}\\{\scriptsize Opus 4.6}} & \shortstack[c]{\includegraphics[height=2.4ex]{logos/gemini-3.1-pro.png}\\{\scriptsize\textbf{Gemini}}\\{\scriptsize 3.1 Pro}} & \shortstack[c]{\includegraphics[height=2.4ex]{logos/kimi-k2.6.png}\\{\scriptsize\textbf{Kimi}}\\{\scriptsize K2.6}} & \shortstack[c]{\includegraphics[height=2.4ex]{logos/mimo-v2.5-pro.png}\\{\scriptsize\textbf{MiMo}}\\{\scriptsize V2.5 Pro}} & \shortstack[c]{\includegraphics[height=2.4ex]{logos/glm-5.png}\\{\scriptsize\textbf{GLM}}\\{\scriptsize 5}} & \shortstack[c]{\includegraphics[height=2.4ex]{logos/deepseek-v4-pro.png}\\{\scriptsize\textbf{DeepSeek}}\\{\scriptsize V4 Pro}} & \shortstack[c]{\includegraphics[height=2.4ex]{logos/gpt-5.4.png}\\{\scriptsize\textbf{GPT}}\\{\scriptsize 5.4}} & \shortstack[c]{\includegraphics[height=2.4ex]{logos/grok-4-20.png}\\{\scriptsize\textbf{Grok}}\\{\scriptsize 4-20}} & \shortstack[c]{\includegraphics[height=2.4ex]{logos/hy3-preview.png}\\{\scriptsize\textbf{Hunyuan}}\\{\scriptsize 3 Preview}} & \shortstack[c]{\includegraphics[height=2.4ex]{logos/minimax-m2.7.png}\\{\scriptsize\textbf{MiniMax}}\\{\scriptsize M2.7}} & \shortstack[c]{\includegraphics[height=2.4ex]{logos/qwen3.6-plus.png}\\{\scriptsize\textbf{Qwen}}\\{\scriptsize 3.6 Plus}} \\
\midrule
\multicolumn{12}{l}{\textit{System Optimization (15 tasks)}} \\
\midrule
{\scriptsize\texttt{aes128\_ctr}} & \cellcolor[HTML]{D2DFC5} \textbf{0.75} & \cellcolor[HTML]{DCE7D2} 0.69 & \cellcolor[HTML]{DAE5CF} \underline{0.70} & \cellcolor[HTML]{E5EDDD} 0.65 & \cellcolor[HTML]{DBE6D0} 0.70 & \cellcolor[HTML]{F9EEF0} 0.37 & \cellcolor[HTML]{F6F9F4} 0.55 & 0.52 & \cellcolor[HTML]{E0EAD7} 0.67 & \cellcolor[HTML]{E7BEC8} 0.00 & \cellcolor[HTML]{E9C2CB} 0.03 \\
{\scriptsize\texttt{agent\_tool\_routing}} & \cellcolor[HTML]{E4ECDC} \textbf{0.65} & \cellcolor[HTML]{FCF6F7} 0.43 & \cellcolor[HTML]{F9EEF0} 0.37 & \cellcolor[HTML]{F5E4E8} 0.30 & \cellcolor[HTML]{FDF8F9} \underline{0.45} & \cellcolor[HTML]{F3DFE4} 0.25 & \cellcolor[HTML]{EDCED6} 0.13 & \cellcolor[HTML]{F2DBE0} 0.22 & \cellcolor[HTML]{EFD2D9} 0.15 & \cellcolor[HTML]{F8EDEF} 0.36 & \cellcolor[HTML]{FAF0F2} 0.38 \\
{\scriptsize\texttt{bm25\_search\_go}} & \cellcolor[HTML]{C4D5B2} \textbf{0.83} & \cellcolor[HTML]{F8FAF6} 0.54 & \cellcolor[HTML]{E6EEDF} \underline{0.64} & 0.51 & 0.51 & \cellcolor[HTML]{EEF3E9} 0.60 & 0.50 & 0.51 & \cellcolor[HTML]{F6E7EB} 0.32 & 0.50 & \cellcolor[HTML]{F4F7F0} 0.56 \\
{\scriptsize\texttt{bvh\_raytracer}} & \cellcolor[HTML]{FCF7F9} \textbf{0.44} & \cellcolor[HTML]{FAF1F3} 0.39 & \cellcolor[HTML]{FBF4F5} \underline{0.41} & \cellcolor[HTML]{FAF2F4} 0.40 & \cellcolor[HTML]{F5E3E7} 0.28 & \cellcolor[HTML]{F9EEF1} 0.37 & \cellcolor[HTML]{EECFD7} 0.13 & \cellcolor[HTML]{F8EDF0} 0.36 & \cellcolor[HTML]{F9EEF1} 0.37 & \cellcolor[HTML]{EBC7D0} 0.07 & \cellcolor[HTML]{ECCCD4} 0.11 \\
{\scriptsize\texttt{concurrent\_kv\_wal}} & \cellcolor[HTML]{ACC593} \textbf{0.96} & \cellcolor[HTML]{D1DFC3} 0.76 & \cellcolor[HTML]{F4F7F0} 0.56 & \cellcolor[HTML]{FEFCFC} 0.47 & \cellcolor[HTML]{E8EFE1} 0.63 & \cellcolor[HTML]{C3D5B1} \underline{0.83} & \cellcolor[HTML]{F9FBF7} 0.53 & \cellcolor[HTML]{F9FBF8} 0.53 & \cellcolor[HTML]{F7E8EB} 0.32 & \cellcolor[HTML]{F4E2E7} 0.28 & \cellcolor[HTML]{F3DFE4} 0.25 \\
{\scriptsize\texttt{fft\_rust}} & \cellcolor[HTML]{E7BEC8} 0.00 & \cellcolor[HTML]{F5F8F2} \underline{0.55} & \cellcolor[HTML]{FAF1F3} 0.39 & \cellcolor[HTML]{F8EDF0} 0.36 & \cellcolor[HTML]{F9FBF7} 0.53 & \cellcolor[HTML]{F2F6EF} \textbf{0.57} & 0.52 & \cellcolor[HTML]{FAFCF9} 0.53 & \cellcolor[HTML]{F9EFF1} 0.37 & \cellcolor[HTML]{F8EBEE} 0.35 & \cellcolor[HTML]{F6F8F3} 0.55 \\
{\scriptsize\texttt{flash\_attention}} & \cellcolor[HTML]{C0D3AD} \textbf{0.85} & \cellcolor[HTML]{FAF0F3} 0.39 & \cellcolor[HTML]{E3ECDB} \underline{0.65} & \cellcolor[HTML]{FCF6F7} 0.43 & \cellcolor[HTML]{F2F6EE} 0.57 & \cellcolor[HTML]{F7E9ED} 0.33 & \cellcolor[HTML]{F7E8EB} 0.32 & \cellcolor[HTML]{FCF7F8} 0.44 & \cellcolor[HTML]{FEFCFD} 0.48 & \cellcolor[HTML]{FCF6F7} 0.43 & \cellcolor[HTML]{F8EBEE} 0.35 \\
{\scriptsize\texttt{gaussian\_blur}} & \cellcolor[HTML]{DAE5CF} \textbf{0.71} & 0.49 & \cellcolor[HTML]{E6EDDE} \underline{0.64} & \cellcolor[HTML]{F7FAF5} 0.54 & \cellcolor[HTML]{F1D8DE} 0.20 & 0.49 & \cellcolor[HTML]{F8EBEE} 0.34 & \cellcolor[HTML]{F5E2E7} 0.28 & \cellcolor[HTML]{F5E2E7} 0.28 & \cellcolor[HTML]{EBC9D2} 0.09 & \cellcolor[HTML]{E7BEC8} 0.00 \\
{\scriptsize\texttt{hash\_join}} & \cellcolor[HTML]{C7D8B6} \textbf{0.81} & \cellcolor[HTML]{E2EBDA} 0.66 & \cellcolor[HTML]{DCE6D1} 0.70 & \cellcolor[HTML]{E2EBD9} 0.66 & \cellcolor[HTML]{DFE8D5} 0.68 & \cellcolor[HTML]{DCE6D1} \underline{0.70} & \cellcolor[HTML]{EBF1E5} 0.61 & \cellcolor[HTML]{F0F5EC} 0.58 & \cellcolor[HTML]{E3EBDB} 0.65 & \cellcolor[HTML]{F0F5EC} 0.58 & \cellcolor[HTML]{E1EAD8} 0.67 \\
{\scriptsize\texttt{levenshtein\_distance}} & \cellcolor[HTML]{F1F5ED} \textbf{0.58} & \cellcolor[HTML]{F9FBF7} \underline{0.53} & 0.52 & \cellcolor[HTML]{EBC8D0} 0.08 & \cellcolor[HTML]{EDCDD5} 0.12 & 0.49 & \cellcolor[HTML]{FEFCFC} 0.47 & \cellcolor[HTML]{E8BFC8} 0.00 & \cellcolor[HTML]{E9C3CC} 0.04 & \cellcolor[HTML]{EBC9D1} 0.08 & \cellcolor[HTML]{E8BFC9} 0.01 \\
{\scriptsize\texttt{radix\_sort}} & \cellcolor[HTML]{D8E4CD} \textbf{0.71} & \cellcolor[HTML]{E5EDDE} 0.64 & \cellcolor[HTML]{FDFAFA} 0.46 & \cellcolor[HTML]{E1EAD8} 0.67 & \cellcolor[HTML]{E9EFE2} 0.62 & \cellcolor[HTML]{DCE6D2} \underline{0.69} & \cellcolor[HTML]{F9FBF7} 0.54 & \cellcolor[HTML]{E8EFE1} 0.63 & \cellcolor[HTML]{E1EAD9} 0.66 & \cellcolor[HTML]{F2DBE0} 0.22 & \cellcolor[HTML]{DFE9D6} 0.68 \\
{\scriptsize\texttt{regex\_engine}} & \cellcolor[HTML]{A5C08A} \textbf{1.00} & \cellcolor[HTML]{FBF3F5} \underline{0.41} & \cellcolor[HTML]{F9EEF1} 0.37 & \cellcolor[HTML]{E8C0CA} 0.02 & \cellcolor[HTML]{EDCDD5} 0.12 & \cellcolor[HTML]{E7BEC8} 0.00 & \cellcolor[HTML]{F0D7DD} 0.19 & \cellcolor[HTML]{EBC9D1} 0.08 & \cellcolor[HTML]{EBC7CF} 0.07 & \cellcolor[HTML]{EBC9D1} 0.08 & \cellcolor[HTML]{E7BEC8} 0.00 \\
{\scriptsize\texttt{sha256\_throughput}} & \cellcolor[HTML]{FDF9FA} \textbf{0.46} & \cellcolor[HTML]{F8EBEE} 0.35 & \cellcolor[HTML]{FCF7F8} \underline{0.44} & \cellcolor[HTML]{F4E0E4} 0.26 & \cellcolor[HTML]{F0D5DC} 0.18 & \cellcolor[HTML]{EED0D7} 0.13 & \cellcolor[HTML]{EED0D7} 0.14 & \cellcolor[HTML]{EED1D8} 0.15 & \cellcolor[HTML]{ECCAD2} 0.09 & \cellcolor[HTML]{EED0D7} 0.14 & \cellcolor[HTML]{EECFD7} 0.13 \\
{\scriptsize\texttt{sstable\_compaction\_rs}} & \cellcolor[HTML]{C4D5B2} \textbf{0.83} & \cellcolor[HTML]{F4E1E5} 0.27 & \cellcolor[HTML]{DAE5CF} 0.70 & \cellcolor[HTML]{FAF2F4} 0.40 & \cellcolor[HTML]{ECF2E6} 0.61 & \cellcolor[HTML]{D7E3CC} \underline{0.72} & \cellcolor[HTML]{F0D5DC} 0.18 & \cellcolor[HTML]{EFF4EA} 0.59 & \cellcolor[HTML]{EAF0E4} 0.62 & \cellcolor[HTML]{FAF1F3} 0.39 & 0.52 \\
{\scriptsize\texttt{z\_order\_range\_scan}} & \cellcolor[HTML]{FDF8F9} \textbf{0.45} & \cellcolor[HTML]{F1D7DD} 0.20 & \cellcolor[HTML]{F6E6EA} 0.31 & \cellcolor[HTML]{F0D5DB} 0.17 & \cellcolor[HTML]{FBF4F5} \underline{0.41} & \cellcolor[HTML]{F6E6EA} 0.31 & \cellcolor[HTML]{F6E7EB} 0.32 & \cellcolor[HTML]{F4E0E5} 0.26 & \cellcolor[HTML]{F8EBEE} 0.35 & \cellcolor[HTML]{EED2D9} 0.15 & \cellcolor[HTML]{ECCBD3} 0.10 \\
\midrule
\multicolumn{12}{l}{\textit{Puzzle \& Challenge (10 tasks)}} \\
\midrule
{\scriptsize\texttt{adaptive\_compression}} & \cellcolor[HTML]{DFE8D5} \textbf{0.68} & \cellcolor[HTML]{F8FAF6} \underline{0.54} & \cellcolor[HTML]{F9EEF1} 0.37 & \cellcolor[HTML]{F4E2E6} 0.28 & \cellcolor[HTML]{F2DCE2} 0.23 & \cellcolor[HTML]{F6E6EA} 0.31 & \cellcolor[HTML]{FCF6F8} 0.43 & \cellcolor[HTML]{EED0D7} 0.14 & \cellcolor[HTML]{E7BEC8} 0.00 & \cellcolor[HTML]{F2DCE1} 0.23 & 0.49 \\
{\scriptsize\texttt{adversarial\_splay}} & \cellcolor[HTML]{ECF2E6} \textbf{0.61} & \cellcolor[HTML]{F8FAF7} 0.54 & \cellcolor[HTML]{F3F7F0} 0.56 & 0.51 & \cellcolor[HTML]{F7F9F4} 0.55 & \cellcolor[HTML]{E7BEC8} 0.00 & \cellcolor[HTML]{F8ECEF} 0.36 & \cellcolor[HTML]{EFD4DA} 0.17 & \cellcolor[HTML]{F0F5EC} \underline{0.58} & \cellcolor[HTML]{F8EBEE} 0.35 & \cellcolor[HTML]{F8EBEE} 0.35 \\
{\scriptsize\texttt{discover\_sorting}} & \cellcolor[HTML]{A5C08A} \textbf{1.00} & \cellcolor[HTML]{AEC696} \underline{0.95} & \cellcolor[HTML]{B7CCA1} 0.90 & \cellcolor[HTML]{B7CCA1} 0.90 & \cellcolor[HTML]{C0D3AD} 0.85 & \cellcolor[HTML]{F3F7EF} 0.57 & \cellcolor[HTML]{C0D3AD} 0.85 & \cellcolor[HTML]{C0D3AD} 0.85 & \cellcolor[HTML]{F3DFE4} 0.25 & \cellcolor[HTML]{E7BEC8} 0.00 & \cellcolor[HTML]{F3F7EF} 0.57 \\
{\scriptsize\texttt{fredkin\_sort\_network}} & \cellcolor[HTML]{ADC594} \textbf{0.96} & \cellcolor[HTML]{F6E6EA} 0.31 & \cellcolor[HTML]{FDFBFB} 0.47 & \cellcolor[HTML]{F1DADF} 0.21 & \cellcolor[HTML]{EDF2E8} 0.60 & \cellcolor[HTML]{F5E4E8} 0.29 & \cellcolor[HTML]{F7E9ED} 0.33 & \cellcolor[HTML]{E9F0E2} \underline{0.62} & \cellcolor[HTML]{F8ECEF} 0.36 & \cellcolor[HTML]{E7BEC8} 0.00 & \cellcolor[HTML]{E7BEC8} 0.00 \\
{\scriptsize\texttt{resnet\_bit\_flip}} & \cellcolor[HTML]{E7EEE0} 0.63 & \cellcolor[HTML]{B6CCA0} \underline{0.90} & \cellcolor[HTML]{F7E8EB} 0.32 & \cellcolor[HTML]{B4CA9D} \textbf{0.92} & \cellcolor[HTML]{EAC6CF} 0.06 & \cellcolor[HTML]{EEF3E9} 0.60 & \cellcolor[HTML]{F7E8EB} 0.32 & \cellcolor[HTML]{F6E7EB} 0.32 & \cellcolor[HTML]{F4E0E4} 0.26 & \cellcolor[HTML]{F8EBEE} 0.35 & \cellcolor[HTML]{FAFCF9} 0.53 \\
{\scriptsize\texttt{safety\_router}} & \cellcolor[HTML]{A6C18C} \textbf{0.99} & \cellcolor[HTML]{A6C18C} \underline{0.99} & \cellcolor[HTML]{E2EBD9} 0.66 & \cellcolor[HTML]{ABC492} 0.96 & \cellcolor[HTML]{E6EDDE} 0.64 & \cellcolor[HTML]{E2EBD9} 0.66 & \cellcolor[HTML]{E7BEC8} 0.00 & \cellcolor[HTML]{F6E7EA} 0.31 & \cellcolor[HTML]{E7BEC8} 0.00 & \cellcolor[HTML]{F6E6EA} 0.31 & \cellcolor[HTML]{E2EBD9} 0.66 \\
{\scriptsize\texttt{smallest\_game\_player}} & \cellcolor[HTML]{EBF1E5} \textbf{0.61} & \cellcolor[HTML]{E7BEC8} 0.00 & \cellcolor[HTML]{E7BEC8} 0.00 & \cellcolor[HTML]{ECCAD2} 0.09 & \cellcolor[HTML]{E7BEC8} 0.00 & \cellcolor[HTML]{E7BEC8} 0.00 & \cellcolor[HTML]{F6E7EB} \underline{0.32} & \cellcolor[HTML]{E7BEC8} 0.00 & \cellcolor[HTML]{E7BEC8} 0.00 & \cellcolor[HTML]{E7BEC8} 0.00 & \cellcolor[HTML]{E7BEC8} 0.00 \\
{\scriptsize\texttt{stack\_machine\_golf}} & \cellcolor[HTML]{A5C08A} \textbf{1.00} & \cellcolor[HTML]{A5C08A} \underline{1.00} & \cellcolor[HTML]{E1EAD8} 0.67 & \cellcolor[HTML]{CBDABB} 0.79 & 0.48 & \cellcolor[HTML]{A5C08A} 1.00 & \cellcolor[HTML]{EFD2D9} 0.16 & \cellcolor[HTML]{ACC594} 0.96 & \cellcolor[HTML]{F9EFF1} 0.38 & \cellcolor[HTML]{F0D6DC} 0.19 & \cellcolor[HTML]{E7BEC8} 0.00 \\
{\scriptsize\texttt{toy\_isa\_opt}} & \cellcolor[HTML]{A9C38F} \underline{0.98} & \cellcolor[HTML]{A5C08A} \textbf{1.00} & \cellcolor[HTML]{E7BEC8} 0.00 & \cellcolor[HTML]{C2D4B0} 0.84 & \cellcolor[HTML]{AAC390} 0.97 & \cellcolor[HTML]{F0F4EB} 0.59 & \cellcolor[HTML]{B9CEA4} 0.89 & \cellcolor[HTML]{ABC491} 0.97 & \cellcolor[HTML]{E3ECDB} 0.65 & \cellcolor[HTML]{C8D8B8} 0.80 & \cellcolor[HTML]{E7BEC8} 0.00 \\
{\scriptsize\texttt{vliw\_scheduler}} & \cellcolor[HTML]{A5C08A} \textbf{1.00} & \cellcolor[HTML]{B0C898} 0.94 & \cellcolor[HTML]{C1D3AE} 0.84 & \cellcolor[HTML]{C0D3AD} 0.85 & \cellcolor[HTML]{F9FBF7} 0.53 & \cellcolor[HTML]{F6E7EB} 0.31 & \cellcolor[HTML]{C4D5B2} 0.83 & \cellcolor[HTML]{A7C18D} 0.99 & \cellcolor[HTML]{A5C08A} \underline{1.00} & \cellcolor[HTML]{FAF1F3} 0.39 & \cellcolor[HTML]{E7BEC8} 0.00 \\
\midrule
\multicolumn{12}{l}{\textit{Model Development (7 tasks)}} \\
\midrule
{\scriptsize\texttt{data\_select\_ifeval}} & \cellcolor[HTML]{E5EDDE} \textbf{0.64} & \underline{0.49} & \cellcolor[HTML]{FEFCFC} 0.48 & \cellcolor[HTML]{F9EFF2} 0.38 & \cellcolor[HTML]{F5E2E7} 0.28 & \cellcolor[HTML]{FDFAFB} 0.46 & \cellcolor[HTML]{ECCAD2} 0.09 & 0.49 & \cellcolor[HTML]{F2DCE2} 0.23 & \cellcolor[HTML]{FCF5F7} 0.43 & \cellcolor[HTML]{F5E3E7} 0.28 \\
{\scriptsize\texttt{flux2\_klein\_lora}} & \textbf{0.48} & \cellcolor[HTML]{EBC7CF} 0.07 & \cellcolor[HTML]{E7BEC8} 0.00 & \cellcolor[HTML]{EFD4DB} 0.17 & \cellcolor[HTML]{F2DBE0} 0.22 & \cellcolor[HTML]{E7BEC8} 0.00 & \cellcolor[HTML]{F5E3E7} 0.28 & \cellcolor[HTML]{ECCAD2} 0.09 & \cellcolor[HTML]{E7BEC8} 0.00 & \cellcolor[HTML]{F5E3E8} \underline{0.29} & \cellcolor[HTML]{EAC6CF} 0.06 \\
{\scriptsize\texttt{grpo\_multisource}} & \cellcolor[HTML]{C1D3AE} 0.84 & \cellcolor[HTML]{CADABA} 0.79 & \cellcolor[HTML]{F0F4EB} 0.59 & \cellcolor[HTML]{C8D8B7} 0.81 & \cellcolor[HTML]{C5D6B4} 0.82 & \cellcolor[HTML]{C0D2AD} \underline{0.85} & \cellcolor[HTML]{C4D5B2} 0.83 & \cellcolor[HTML]{E7BEC8} 0.00 & \cellcolor[HTML]{F2F6EF} 0.57 & \cellcolor[HTML]{B1C899} \textbf{0.93} & \cellcolor[HTML]{C1D3AE} 0.84 \\
{\scriptsize\texttt{llm\_online\_serving}} & \cellcolor[HTML]{E7BEC8} 0.00 & \cellcolor[HTML]{E7BEC8} 0.00 & \cellcolor[HTML]{E7BEC8} 0.00 & \cellcolor[HTML]{E8BFC9} 0.01 & \cellcolor[HTML]{E9C2CC} 0.03 & \cellcolor[HTML]{E9C3CC} 0.04 & \cellcolor[HTML]{F6E6EA} \underline{0.31} & \cellcolor[HTML]{F7EAED} \textbf{0.34} & \cellcolor[HTML]{E7BEC8} 0.00 & \cellcolor[HTML]{E7BEC8} 0.00 & \cellcolor[HTML]{F4E2E6} 0.28 \\
{\scriptsize\texttt{moving\_mnist\_world\_model}} & \cellcolor[HTML]{DEE8D4} \textbf{0.68} & \cellcolor[HTML]{F6E7EB} \underline{0.32} & \cellcolor[HTML]{F2DAE0} 0.22 & \cellcolor[HTML]{F5E5E9} 0.30 & \cellcolor[HTML]{F6E6E9} 0.30 & \cellcolor[HTML]{F1D7DD} 0.19 & \cellcolor[HTML]{EFD3DA} 0.16 & \cellcolor[HTML]{E8C1CB} 0.02 & \cellcolor[HTML]{F4E2E6} 0.28 & \cellcolor[HTML]{F0D7DD} 0.19 & \cellcolor[HTML]{F0D6DC} 0.18 \\
{\scriptsize\texttt{multilingual\_ocr}} & \cellcolor[HTML]{B9CEA5} \underline{0.89} & \cellcolor[HTML]{DEE8D4} 0.69 & \cellcolor[HTML]{A8C28E} \textbf{0.98} & \cellcolor[HTML]{BBCFA6} 0.88 & \cellcolor[HTML]{BBCFA6} 0.88 & \cellcolor[HTML]{E8EFE1} 0.63 & \cellcolor[HTML]{FCF7F9} 0.44 & \cellcolor[HTML]{E7BEC8} 0.00 & \cellcolor[HTML]{C4D6B2} 0.83 & \cellcolor[HTML]{DBE6D0} 0.70 & \cellcolor[HTML]{FDF9FA} 0.45 \\
{\scriptsize\texttt{scaling\_law}} & \cellcolor[HTML]{C0D3AD} \textbf{0.85} & \cellcolor[HTML]{EED0D7} 0.14 & \cellcolor[HTML]{E4ECDC} 0.65 & \cellcolor[HTML]{CCDBBC} \underline{0.78} & \cellcolor[HTML]{ECF2E6} 0.61 & \cellcolor[HTML]{F6E5E9} 0.30 & \cellcolor[HTML]{F7E9ED} 0.33 & \cellcolor[HTML]{E7BEC8} 0.00 & \cellcolor[HTML]{E7BEC8} 0.00 & \cellcolor[HTML]{FCF6F7} 0.43 & \cellcolor[HTML]{E8EFE1} 0.63 \\
\midrule
\multicolumn{12}{l}{\textit{CUDA (4 tasks)}} \\
\midrule
{\scriptsize\texttt{huffman\_canonical\_decode}} & \cellcolor[HTML]{FDF9FA} \textbf{0.45} & \cellcolor[HTML]{F9EFF2} \underline{0.38} & \cellcolor[HTML]{F1D8DE} 0.20 & \cellcolor[HTML]{ECCCD4} 0.11 & \cellcolor[HTML]{ECCCD4} 0.11 & \cellcolor[HTML]{E7BEC8} 0.00 & \cellcolor[HTML]{E7BEC8} 0.00 & \cellcolor[HTML]{EBC9D2} 0.09 & \cellcolor[HTML]{E8C1CB} 0.02 & \cellcolor[HTML]{E7BEC8} 0.00 & \cellcolor[HTML]{E7BEC8} 0.00 \\
{\scriptsize\texttt{icp\_correspondence\_step}} & \cellcolor[HTML]{F6F9F4} \textbf{0.55} & \underline{0.52} & 0.51 & \cellcolor[HTML]{FDF8F9} 0.45 & 0.50 & \cellcolor[HTML]{E7BEC8} 0.00 & \cellcolor[HTML]{FCF7F8} 0.44 & \cellcolor[HTML]{F5E3E7} 0.28 & \cellcolor[HTML]{EED2D8} 0.15 & \cellcolor[HTML]{F4E1E6} 0.27 & \cellcolor[HTML]{E7BEC8} 0.00 \\
{\scriptsize\texttt{msm\_pippenger\_bls12\_381}} & \cellcolor[HTML]{EFD3DA} \underline{0.16} & \cellcolor[HTML]{E7BEC8} 0.00 & \cellcolor[HTML]{ECCBD3} 0.10 & \cellcolor[HTML]{E7BEC8} 0.00 & \cellcolor[HTML]{F2DAE0} \textbf{0.21} & \cellcolor[HTML]{E7BEC8} 0.00 & \cellcolor[HTML]{ECCCD3} 0.10 & \cellcolor[HTML]{ECCCD3} 0.10 & \cellcolor[HTML]{E7BEC8} 0.00 & \cellcolor[HTML]{E7BEC8} 0.00 & \cellcolor[HTML]{E7BEC8} 0.00 \\
{\scriptsize\texttt{ntt\_butterfly}} & \cellcolor[HTML]{F9EFF1} \textbf{0.37} & \cellcolor[HTML]{E7BEC8} 0.00 & \cellcolor[HTML]{EFD4DB} 0.17 & \cellcolor[HTML]{EFD2D9} 0.15 & \cellcolor[HTML]{EFD2D9} 0.15 & \cellcolor[HTML]{E7BEC8} 0.00 & \cellcolor[HTML]{E7BEC8} 0.00 & \cellcolor[HTML]{F3DEE3} \underline{0.25} & \cellcolor[HTML]{ECCCD4} 0.11 & \cellcolor[HTML]{F2DBE0} 0.22 & \cellcolor[HTML]{E7BEC8} 0.00 \\
\bottomrule
\end{tabular}%
}
\caption{Per-task Avg@3 results. Per-row best is \textbf{bold} and runner-up \underline{underlined}.}
\label{tab:per_task}
\end{table*}

\begin{table*}[h]
\centering
\footnotesize
\setlength{\tabcolsep}{1pt}
\renewcommand{\arraystretch}{1.15}
\setlength{\aboverulesep}{1pt}
\setlength{\belowrulesep}{1pt}
\resizebox{\textwidth}{!}{%
\begin{tabular}{@{}p{0.22\textwidth}>{\centering\arraybackslash}p{0.0673\textwidth}>{\centering\arraybackslash}p{0.0673\textwidth}>{\centering\arraybackslash}p{0.0673\textwidth}>{\centering\arraybackslash}p{0.0673\textwidth}>{\centering\arraybackslash}p{0.0673\textwidth}>{\centering\arraybackslash}p{0.0673\textwidth}>{\centering\arraybackslash}p{0.0673\textwidth}>{\centering\arraybackslash}p{0.0673\textwidth}>{\centering\arraybackslash}p{0.0673\textwidth}>{\centering\arraybackslash}p{0.0673\textwidth}>{\centering\arraybackslash}p{0.0673\textwidth}@{}}
\toprule
\textbf{Task} & \shortstack[c]{\includegraphics[height=2.4ex]{logos/claude-opus-4.6.png}\\{\scriptsize\textbf{Claude}}\\{\scriptsize Opus 4.6}} & \shortstack[c]{\includegraphics[height=2.4ex]{logos/gemini-3.1-pro.png}\\{\scriptsize\textbf{Gemini}}\\{\scriptsize 3.1 Pro}} & \shortstack[c]{\includegraphics[height=2.4ex]{logos/kimi-k2.6.png}\\{\scriptsize\textbf{Kimi}}\\{\scriptsize K2.6}} & \shortstack[c]{\includegraphics[height=2.4ex]{logos/mimo-v2.5-pro.png}\\{\scriptsize\textbf{MiMo}}\\{\scriptsize V2.5 Pro}} & \shortstack[c]{\includegraphics[height=2.4ex]{logos/glm-5.png}\\{\scriptsize\textbf{GLM}}\\{\scriptsize 5}} & \shortstack[c]{\includegraphics[height=2.4ex]{logos/deepseek-v4-pro.png}\\{\scriptsize\textbf{DeepSeek}}\\{\scriptsize V4 Pro}} & \shortstack[c]{\includegraphics[height=2.4ex]{logos/gpt-5.4.png}\\{\scriptsize\textbf{GPT}}\\{\scriptsize 5.4}} & \shortstack[c]{\includegraphics[height=2.4ex]{logos/grok-4-20.png}\\{\scriptsize\textbf{Grok}}\\{\scriptsize 4-20}} & \shortstack[c]{\includegraphics[height=2.4ex]{logos/hy3-preview.png}\\{\scriptsize\textbf{Hunyuan}}\\{\scriptsize 3 Preview}} & \shortstack[c]{\includegraphics[height=2.4ex]{logos/minimax-m2.7.png}\\{\scriptsize\textbf{MiniMax}}\\{\scriptsize M2.7}} & \shortstack[c]{\includegraphics[height=2.4ex]{logos/qwen3.6-plus.png}\\{\scriptsize\textbf{Qwen}}\\{\scriptsize 3.6 Plus}} \\
\midrule
\multicolumn{12}{l}{\textit{System Optimization (15 tasks)}} \\
\midrule
{\scriptsize\texttt{aes128\_ctr}} & \cellcolor[HTML]{CCDBBD} \textbf{0.78} & \cellcolor[HTML]{DBE5D0} 0.70 & \cellcolor[HTML]{DAE5CF} 0.70 & \cellcolor[HTML]{DAE5CF} 0.70 & \cellcolor[HTML]{D1DFC4} \underline{0.75} & \cellcolor[HTML]{EDF2E8} 0.60 & \cellcolor[HTML]{F3F7EF} 0.57 & \cellcolor[HTML]{DBE6D0} 0.70 & \cellcolor[HTML]{D9E4CE} 0.71 & \cellcolor[HTML]{E7BEC8} 0.00 & \cellcolor[HTML]{E9C2CB} 0.03 \\
{\scriptsize\texttt{agent\_tool\_routing}} & \cellcolor[HTML]{D9E4CD} \textbf{0.71} & 0.50 & 0.49 & \cellcolor[HTML]{F8EBEE} 0.35 & \underline{0.51} & \cellcolor[HTML]{F7E8EB} 0.32 & \cellcolor[HTML]{EFD3D9} 0.16 & \cellcolor[HTML]{F8ECEF} 0.36 & \cellcolor[HTML]{F2DCE1} 0.23 & \cellcolor[HTML]{FAF1F3} 0.39 & \cellcolor[HTML]{FEFBFC} 0.47 \\
{\scriptsize\texttt{bm25\_search\_go}} & \cellcolor[HTML]{A5C08A} \textbf{1.00} & \cellcolor[HTML]{EEF3E9} 0.59 & \cellcolor[HTML]{E3ECDB} \underline{0.65} & \cellcolor[HTML]{FAFBF8} 0.53 & 0.52 & \cellcolor[HTML]{E6EDDE} 0.64 & 0.51 & 0.52 & 0.51 & 0.51 & \cellcolor[HTML]{F1F5ED} 0.58 \\
{\scriptsize\texttt{bvh\_raytracer}} & \cellcolor[HTML]{FDFBFB} \textbf{0.47} & \cellcolor[HTML]{FAF2F4} 0.40 & \cellcolor[HTML]{FCF6F7} 0.43 & \cellcolor[HTML]{FBF5F7} 0.42 & \cellcolor[HTML]{FCF6F8} \underline{0.43} & \cellcolor[HTML]{FBF5F6} 0.42 & \cellcolor[HTML]{FAF2F4} 0.40 & \cellcolor[HTML]{FAF2F4} 0.40 & \cellcolor[HTML]{FBF3F5} 0.41 & \cellcolor[HTML]{F2DAE0} 0.22 & \cellcolor[HTML]{F6E7EA} 0.31 \\
{\scriptsize\texttt{concurrent\_kv\_wal}} & \cellcolor[HTML]{ABC492} \underline{0.96} & \cellcolor[HTML]{CEDCBF} 0.77 & \cellcolor[HTML]{B3C99C} 0.92 & \cellcolor[HTML]{CFDEC1} 0.76 & \cellcolor[HTML]{A5C08A} \textbf{1.00} & \cellcolor[HTML]{B1C89A} 0.93 & \cellcolor[HTML]{BED1AB} 0.86 & \cellcolor[HTML]{DCE6D1} 0.70 & \cellcolor[HTML]{ACC593} 0.96 & \cellcolor[HTML]{C3D5B1} 0.83 & \cellcolor[HTML]{D2DFC4} 0.75 \\
{\scriptsize\texttt{fft\_rust}} & \cellcolor[HTML]{E7BEC8} 0.00 & \cellcolor[HTML]{ECF2E7} \textbf{0.60} & \cellcolor[HTML]{EFF4EA} \underline{0.59} & \cellcolor[HTML]{F5F8F2} 0.55 & \cellcolor[HTML]{F7FAF5} 0.54 & \cellcolor[HTML]{F0F4EB} 0.58 & \cellcolor[HTML]{FBFCF9} 0.52 & \cellcolor[HTML]{FAFCF9} 0.53 & \cellcolor[HTML]{EFF4EA} 0.59 & \cellcolor[HTML]{FAFBF8} 0.53 & \cellcolor[HTML]{F1F5ED} 0.58 \\
{\scriptsize\texttt{flash\_attention}} & \cellcolor[HTML]{AFC797} \textbf{0.94} & \cellcolor[HTML]{DCE6D2} 0.69 & \cellcolor[HTML]{DBE6D1} \underline{0.70} & \cellcolor[HTML]{F1F5ED} 0.58 & \cellcolor[HTML]{EBF1E5} 0.61 & \cellcolor[HTML]{FAFBF8} 0.53 & 0.52 & \cellcolor[HTML]{FCF7F8} 0.44 & 0.51 & 0.51 & \cellcolor[HTML]{F6F8F3} 0.55 \\
{\scriptsize\texttt{gaussian\_blur}} & \cellcolor[HTML]{D4E1C7} \textbf{0.74} & \cellcolor[HTML]{E6EDDE} 0.64 & \cellcolor[HTML]{E3EBDB} \underline{0.65} & \cellcolor[HTML]{EBF1E5} 0.61 & \cellcolor[HTML]{EDF3E8} 0.60 & \cellcolor[HTML]{F1F5ED} 0.58 & \cellcolor[HTML]{F8EDF0} 0.36 & \cellcolor[HTML]{F5E3E7} 0.28 & \cellcolor[HTML]{F6E6EA} 0.31 & \cellcolor[HTML]{F4E0E5} 0.26 & \cellcolor[HTML]{E7BEC8} 0.00 \\
{\scriptsize\texttt{hash\_join}} & \cellcolor[HTML]{A5C08A} \textbf{1.00} & \cellcolor[HTML]{DCE6D1} 0.70 & \cellcolor[HTML]{D8E4CD} \underline{0.71} & \cellcolor[HTML]{DCE6D1} 0.69 & \cellcolor[HTML]{DCE6D1} 0.70 & \cellcolor[HTML]{DBE5D0} 0.70 & \cellcolor[HTML]{E2EAD9} 0.66 & \cellcolor[HTML]{EDF3E8} 0.60 & \cellcolor[HTML]{DDE7D2} 0.69 & \cellcolor[HTML]{EFF4EA} 0.59 & \cellcolor[HTML]{D9E5CE} 0.71 \\
{\scriptsize\texttt{levenshtein\_distance}} & \cellcolor[HTML]{EFF4EA} \textbf{0.59} & \cellcolor[HTML]{F2F6EE} \underline{0.57} & \cellcolor[HTML]{F5F8F2} 0.56 & \cellcolor[HTML]{EDCFD6} 0.13 & \cellcolor[HTML]{EDCED6} 0.12 & 0.50 & \cellcolor[HTML]{FEFCFD} 0.48 & \cellcolor[HTML]{E8BFC9} 0.01 & \cellcolor[HTML]{EDCDD5} 0.12 & \cellcolor[HTML]{EDCFD6} 0.13 & \cellcolor[HTML]{E8BFC9} 0.01 \\
{\scriptsize\texttt{radix\_sort}} & \cellcolor[HTML]{D8E4CC} \textbf{0.72} & \cellcolor[HTML]{DEE8D5} 0.68 & \cellcolor[HTML]{DDE7D3} 0.69 & \cellcolor[HTML]{DEE8D4} 0.68 & \cellcolor[HTML]{E4ECDC} 0.65 & \cellcolor[HTML]{DBE6D1} \underline{0.70} & \cellcolor[HTML]{F2F6EE} 0.57 & \cellcolor[HTML]{E4ECDC} 0.65 & \cellcolor[HTML]{DCE6D1} 0.70 & \cellcolor[HTML]{E1EAD9} 0.66 & \cellcolor[HTML]{DEE8D4} 0.68 \\
{\scriptsize\texttt{regex\_engine}} & \cellcolor[HTML]{A5C08A} \textbf{1.00} & \cellcolor[HTML]{FCF6F7} 0.43 & \cellcolor[HTML]{BBCFA7} \underline{0.88} & \cellcolor[HTML]{EAC4CD} 0.05 & \cellcolor[HTML]{EED1D8} 0.14 & \cellcolor[HTML]{E7BEC8} 0.00 & \cellcolor[HTML]{F1F5EC} 0.58 & \cellcolor[HTML]{F3DFE4} 0.25 & \cellcolor[HTML]{ECCCD4} 0.10 & \cellcolor[HTML]{F2DAE0} 0.21 & \cellcolor[HTML]{E7BEC8} 0.00 \\
{\scriptsize\texttt{sha256\_throughput}} & \cellcolor[HTML]{FDF9FA} \textbf{0.46} & \cellcolor[HTML]{FDF9FA} 0.46 & \cellcolor[HTML]{FDF9FA} 0.46 & \cellcolor[HTML]{FDF9FA} \underline{0.46} & \cellcolor[HTML]{F4E0E5} 0.26 & \cellcolor[HTML]{EED0D7} 0.13 & \cellcolor[HTML]{FBF4F6} 0.42 & \cellcolor[HTML]{EFD4DA} 0.17 & \cellcolor[HTML]{EED0D7} 0.14 & \cellcolor[HTML]{EED0D8} 0.14 & \cellcolor[HTML]{EECFD7} 0.13 \\
{\scriptsize\texttt{sstable\_compaction\_rs}} & \cellcolor[HTML]{C2D4B0} \textbf{0.84} & \cellcolor[HTML]{D4E1C7} 0.74 & \cellcolor[HTML]{CEDCBF} 0.77 & \cellcolor[HTML]{EEF3E9} 0.59 & \cellcolor[HTML]{D5E2C9} 0.73 & \cellcolor[HTML]{CEDCBF} 0.77 & \cellcolor[HTML]{F8FAF6} 0.54 & \cellcolor[HTML]{EEF3EA} 0.59 & \cellcolor[HTML]{E1EAD8} 0.67 & \cellcolor[HTML]{E5EDDD} 0.65 & \cellcolor[HTML]{C7D8B6} \underline{0.81} \\
{\scriptsize\texttt{z\_order\_range\_scan}} & \textbf{0.49} & \cellcolor[HTML]{F6E6EA} 0.31 & \cellcolor[HTML]{FBF4F5} 0.41 & \cellcolor[HTML]{F5E3E7} 0.28 & \cellcolor[HTML]{FCF6F7} \underline{0.43} & \cellcolor[HTML]{F8EDF0} 0.36 & \cellcolor[HTML]{FAF0F3} 0.39 & \cellcolor[HTML]{F6E6E9} 0.30 & \cellcolor[HTML]{FBF5F6} 0.42 & \cellcolor[HTML]{F3DDE2} 0.24 & \cellcolor[HTML]{F5E5E9} 0.30 \\
\midrule
\multicolumn{12}{l}{\textit{Puzzle \& Challenge (10 tasks)}} \\
\midrule
{\scriptsize\texttt{adaptive\_compression}} & \cellcolor[HTML]{D9E4CE} \textbf{0.71} & \cellcolor[HTML]{DDE7D3} \underline{0.69} & \cellcolor[HTML]{FEFBFC} 0.47 & \cellcolor[HTML]{ECF1E6} 0.61 & \cellcolor[HTML]{F7EAED} 0.34 & \cellcolor[HTML]{FAFCF9} 0.53 & \cellcolor[HTML]{F0F4EB} 0.58 & \cellcolor[HTML]{F4E0E5} 0.26 & \cellcolor[HTML]{E7BEC8} 0.00 & \cellcolor[HTML]{FAF1F3} 0.39 & \cellcolor[HTML]{FBFCFA} 0.52 \\
{\scriptsize\texttt{adversarial\_splay}} & \cellcolor[HTML]{E4ECDC} \textbf{0.65} & \cellcolor[HTML]{F5F8F1} 0.56 & \cellcolor[HTML]{F2F6EF} 0.57 & \cellcolor[HTML]{F4F8F1} 0.56 & \cellcolor[HTML]{F4F7F0} 0.56 & \cellcolor[HTML]{E7BEC8} 0.00 & \cellcolor[HTML]{F3F6EF} 0.57 & 0.50 & \cellcolor[HTML]{EFF4EA} \underline{0.59} & \cellcolor[HTML]{F4F7F1} 0.56 & \cellcolor[HTML]{F6F9F3} 0.55 \\
{\scriptsize\texttt{discover\_sorting}} & \cellcolor[HTML]{A5C08A} \textbf{1.00} & \cellcolor[HTML]{A5C08A} \underline{1.00} & \cellcolor[HTML]{A5C08A} 1.00 & \cellcolor[HTML]{A5C08A} 1.00 & \cellcolor[HTML]{C0D3AD} 0.85 & \cellcolor[HTML]{C0D3AD} 0.85 & \cellcolor[HTML]{C0D3AD} 0.85 & \cellcolor[HTML]{C0D3AD} 0.85 & \cellcolor[HTML]{D2DFC4} 0.75 & \cellcolor[HTML]{E7BEC8} 0.00 & \cellcolor[HTML]{C0D3AD} 0.85 \\
{\scriptsize\texttt{fredkin\_sort\_network}} & \cellcolor[HTML]{A5C08A} \textbf{1.00} & \cellcolor[HTML]{B1C899} \underline{0.93} & \cellcolor[HTML]{D5E1C8} 0.73 & \cellcolor[HTML]{FAF2F4} 0.40 & \cellcolor[HTML]{E1EAD8} 0.67 & \cellcolor[HTML]{FDFBFB} 0.47 & \cellcolor[HTML]{BDD1A9} 0.87 & \cellcolor[HTML]{E1EAD8} 0.67 & \cellcolor[HTML]{F9FBF7} 0.53 & \cellcolor[HTML]{E7BEC8} 0.00 & \cellcolor[HTML]{E7BEC8} 0.00 \\
{\scriptsize\texttt{resnet\_bit\_flip}} & \cellcolor[HTML]{A7C18D} \textbf{0.99} & \cellcolor[HTML]{AEC696} 0.95 & \cellcolor[HTML]{ACC493} 0.96 & \cellcolor[HTML]{A9C390} \underline{0.97} & \cellcolor[HTML]{F0D6DD} 0.19 & \cellcolor[HTML]{A9C390} 0.97 & \cellcolor[HTML]{ACC493} 0.96 & \cellcolor[HTML]{E2EAD9} 0.66 & \cellcolor[HTML]{CDDCBF} 0.78 & \cellcolor[HTML]{F6F9F3} 0.55 & \cellcolor[HTML]{BED1AA} 0.86 \\
{\scriptsize\texttt{safety\_router}} & \cellcolor[HTML]{A6C18C} \textbf{0.99} & \cellcolor[HTML]{A6C18C} \underline{0.99} & \cellcolor[HTML]{A6C18C} 0.99 & \cellcolor[HTML]{A8C28E} 0.98 & \cellcolor[HTML]{A6C18C} 0.99 & \cellcolor[HTML]{A6C18C} 0.99 & \cellcolor[HTML]{E7BEC8} 0.00 & \cellcolor[HTML]{B0C899} 0.94 & \cellcolor[HTML]{E7BEC8} 0.00 & \cellcolor[HTML]{B3CA9C} 0.92 & \cellcolor[HTML]{A6C18C} 0.99 \\
{\scriptsize\texttt{smallest\_game\_player}} & \cellcolor[HTML]{A7C28D} \textbf{0.99} & \cellcolor[HTML]{E7BEC8} 0.00 & \cellcolor[HTML]{E7BEC8} 0.00 & \cellcolor[HTML]{F5E3E7} 0.28 & \cellcolor[HTML]{E7BEC8} 0.00 & \cellcolor[HTML]{E7BEC8} 0.00 & \cellcolor[HTML]{ADC594} \underline{0.96} & \cellcolor[HTML]{E7BEC8} 0.00 & \cellcolor[HTML]{E7BEC8} 0.00 & \cellcolor[HTML]{E7BEC8} 0.00 & \cellcolor[HTML]{E7BEC8} 0.00 \\
{\scriptsize\texttt{stack\_machine\_golf}} & \cellcolor[HTML]{A5C08A} \textbf{1.00} & \cellcolor[HTML]{A5C08A} \underline{1.00} & \cellcolor[HTML]{A5C08A} 1.00 & \cellcolor[HTML]{A5C08A} 1.00 & \cellcolor[HTML]{F6F9F3} 0.55 & \cellcolor[HTML]{A5C08A} 1.00 & \cellcolor[HTML]{F0D6DD} 0.19 & \cellcolor[HTML]{A5C08A} 1.00 & \cellcolor[HTML]{BBD0A7} 0.88 & \cellcolor[HTML]{F9EEF1} 0.37 & \cellcolor[HTML]{E7BEC8} 0.00 \\
{\scriptsize\texttt{toy\_isa\_opt}} & \cellcolor[HTML]{A5C08A} \textbf{1.00} & \cellcolor[HTML]{A5C08A} \underline{1.00} & \cellcolor[HTML]{E7BEC8} 0.00 & \cellcolor[HTML]{A8C28E} 0.98 & \cellcolor[HTML]{A5C08A} 1.00 & \cellcolor[HTML]{B7CCA1} 0.90 & \cellcolor[HTML]{B7CCA1} 0.90 & \cellcolor[HTML]{AAC391} 0.97 & \cellcolor[HTML]{A8C28E} 0.98 & \cellcolor[HTML]{ABC492} 0.97 & \cellcolor[HTML]{E7BEC8} 0.00 \\
{\scriptsize\texttt{vliw\_scheduler}} & \cellcolor[HTML]{A5C08A} \textbf{1.00} & \cellcolor[HTML]{A5C08A} 1.00 & \cellcolor[HTML]{A8C28E} 0.98 & \cellcolor[HTML]{A8C28E} 0.98 & \cellcolor[HTML]{AFC797} 0.94 & \cellcolor[HTML]{AFC797} 0.94 & \cellcolor[HTML]{ACC594} 0.96 & \cellcolor[HTML]{A5C08A} 1.00 & \cellcolor[HTML]{A5C08A} \underline{1.00} & \cellcolor[HTML]{EEF3E8} 0.60 & \cellcolor[HTML]{E7BEC8} 0.00 \\
\midrule
\multicolumn{12}{l}{\textit{Model Development (7 tasks)}} \\
\midrule
{\scriptsize\texttt{data\_select\_ifeval}} & \cellcolor[HTML]{BED2AB} \underline{0.86} & \cellcolor[HTML]{F0F5EC} 0.58 & \cellcolor[HTML]{D6E2C9} 0.73 & \cellcolor[HTML]{E3EBDB} 0.66 & \cellcolor[HTML]{F7F9F5} 0.54 & \cellcolor[HTML]{AEC696} \textbf{0.95} & \cellcolor[HTML]{F1D9DF} 0.21 & \cellcolor[HTML]{CFDDC1} 0.77 & \cellcolor[HTML]{FAF2F4} 0.40 & \cellcolor[HTML]{E3EBDB} 0.66 & \cellcolor[HTML]{F4F7F0} 0.56 \\
{\scriptsize\texttt{flux2\_klein\_lora}} & \cellcolor[HTML]{D7E3CC} \underline{0.72} & \cellcolor[HTML]{F1D9DE} 0.20 & \cellcolor[HTML]{E7BEC8} 0.00 & \cellcolor[HTML]{F9EFF1} 0.37 & \cellcolor[HTML]{E1EAD9} 0.66 & \cellcolor[HTML]{E7BEC8} 0.00 & \cellcolor[HTML]{FBF3F5} 0.41 & \cellcolor[HTML]{EECFD7} 0.13 & \cellcolor[HTML]{E7BEC8} 0.00 & \cellcolor[HTML]{BED1AA} \textbf{0.86} & \cellcolor[HTML]{EDCDD4} 0.11 \\
{\scriptsize\texttt{grpo\_multisource}} & \cellcolor[HTML]{B9CEA4} \underline{0.89} & \cellcolor[HTML]{C5D6B4} 0.82 & \cellcolor[HTML]{B9CEA4} 0.89 & \cellcolor[HTML]{C1D3AE} 0.84 & \cellcolor[HTML]{C1D3AE} 0.84 & \cellcolor[HTML]{BDD1A9} 0.87 & \cellcolor[HTML]{C1D3AE} 0.84 & \cellcolor[HTML]{E7BEC8} 0.00 & \cellcolor[HTML]{BDD1A9} 0.87 & \cellcolor[HTML]{A9C38F} \textbf{0.98} & \cellcolor[HTML]{B9CEA4} 0.89 \\
{\scriptsize\texttt{llm\_online\_serving}} & \cellcolor[HTML]{E7BEC8} 0.00 & \cellcolor[HTML]{E7BEC8} 0.00 & \cellcolor[HTML]{E7BEC8} 0.00 & \cellcolor[HTML]{E8C0CA} 0.02 & \cellcolor[HTML]{ECCBD3} 0.10 & \cellcolor[HTML]{ECCCD4} 0.11 & \cellcolor[HTML]{C8D8B7} \underline{0.81} & 0.49 & \cellcolor[HTML]{E7BEC8} 0.00 & \cellcolor[HTML]{E7BEC8} 0.00 & \cellcolor[HTML]{C4D6B2} \textbf{0.83} \\
{\scriptsize\texttt{moving\_mnist\_world\_model}} & \cellcolor[HTML]{D1DFC4} \textbf{0.75} & \cellcolor[HTML]{FBF3F5} 0.40 & \cellcolor[HTML]{F9EFF2} 0.38 & \cellcolor[HTML]{FDF8F9} \underline{0.45} & \cellcolor[HTML]{FAF0F2} 0.38 & \cellcolor[HTML]{FAF2F4} 0.40 & \cellcolor[HTML]{FAF1F3} 0.39 & \cellcolor[HTML]{EBC7D0} 0.07 & \cellcolor[HTML]{FCF8F9} 0.44 & \cellcolor[HTML]{FAF1F3} 0.39 & \cellcolor[HTML]{F5E3E7} 0.29 \\
{\scriptsize\texttt{multilingual\_ocr}} & \cellcolor[HTML]{ACC593} 0.96 & \cellcolor[HTML]{B1C899} 0.93 & \cellcolor[HTML]{A5C08A} \textbf{1.00} & \cellcolor[HTML]{B3C99C} 0.92 & \cellcolor[HTML]{A5C08A} \underline{1.00} & \cellcolor[HTML]{A5C08A} 1.00 & \cellcolor[HTML]{D9E4CD} 0.71 & \cellcolor[HTML]{E7BEC8} 0.00 & \cellcolor[HTML]{AEC696} 0.95 & \cellcolor[HTML]{B8CDA3} 0.89 & \cellcolor[HTML]{B7CCA1} 0.90 \\
{\scriptsize\texttt{scaling\_law}} & \cellcolor[HTML]{BACEA5} \underline{0.88} & \cellcolor[HTML]{FBF4F6} 0.42 & \cellcolor[HTML]{D9E4CD} 0.71 & \cellcolor[HTML]{A5C08A} \textbf{1.00} & \cellcolor[HTML]{DDE7D2} 0.69 & \cellcolor[HTML]{EEF3E9} 0.59 & \cellcolor[HTML]{FDF9FA} 0.45 & \cellcolor[HTML]{E7BEC8} 0.00 & \cellcolor[HTML]{E7BEC8} 0.00 & \cellcolor[HTML]{E1EAD8} 0.67 & \cellcolor[HTML]{DAE5CE} 0.71 \\
\midrule
\multicolumn{12}{l}{\textit{CUDA (4 tasks)}} \\
\midrule
{\scriptsize\texttt{huffman\_canonical\_decode}} & \textbf{0.48} & \cellcolor[HTML]{FDF9FA} \underline{0.45} & \cellcolor[HTML]{F9EEF1} 0.37 & \cellcolor[HTML]{F6E7EB} 0.32 & \cellcolor[HTML]{F4E1E6} 0.27 & \cellcolor[HTML]{E7BEC8} 0.00 & \cellcolor[HTML]{E7BEC8} 0.00 & \cellcolor[HTML]{EED0D8} 0.14 & \cellcolor[HTML]{EBC7D0} 0.07 & \cellcolor[HTML]{E7BEC8} 0.00 & \cellcolor[HTML]{E7BEC8} 0.00 \\
{\scriptsize\texttt{icp\_correspondence\_step}} & \cellcolor[HTML]{EEF3E8} \textbf{0.60} & \cellcolor[HTML]{F9FBF7} \underline{0.53} & 0.52 & 0.51 & 0.50 & \cellcolor[HTML]{E7BEC8} 0.00 & 0.48 & \cellcolor[HTML]{F7E8EC} 0.33 & \cellcolor[HTML]{FDF8F9} 0.45 & \cellcolor[HTML]{F6E7EA} 0.31 & \cellcolor[HTML]{E7BEC8} 0.00 \\
{\scriptsize\texttt{msm\_pippenger\_bls12\_381}} & \textbf{0.48} & \cellcolor[HTML]{E7BEC8} 0.00 & \cellcolor[HTML]{F6E6EA} 0.31 & \cellcolor[HTML]{E7BEC8} 0.00 & \cellcolor[HTML]{F7E8EC} \underline{0.32} & \cellcolor[HTML]{E7BEC8} 0.00 & \cellcolor[HTML]{F6E7EA} 0.31 & \cellcolor[HTML]{F6E7EA} 0.31 & \cellcolor[HTML]{E7BEC8} 0.00 & \cellcolor[HTML]{E7BEC8} 0.00 & \cellcolor[HTML]{E7BEC8} 0.00 \\
{\scriptsize\texttt{ntt\_butterfly}} & \cellcolor[HTML]{F1F5EC} \textbf{0.58} & \cellcolor[HTML]{E7BEC8} 0.00 & \underline{0.51} & \cellcolor[HTML]{FDF9FA} 0.46 & \cellcolor[HTML]{FDFAFB} 0.46 & \cellcolor[HTML]{E7BEC8} 0.00 & \cellcolor[HTML]{E7BEC8} 0.00 & \cellcolor[HTML]{F7EAED} 0.34 & \cellcolor[HTML]{F6E7EA} 0.31 & \cellcolor[HTML]{FCF8F9} 0.44 & \cellcolor[HTML]{E7BEC8} 0.00 \\
\bottomrule
\end{tabular}%
}
\caption{Per-task Best@3 results. Per-row best is \textbf{bold} and runner-up \underline{underlined}. Columns are in the same order as the Avg@3 table (by overall Avg@3).}
\label{tab:per_task_best}
\end{table*}

\clearpage
\section{More on Analysis}
\label{app:more_analysis}

\subsection{Model Generations}
\label{app:generations}

\begin{figure}[ht]
\centering
\includegraphics[width=\textwidth]{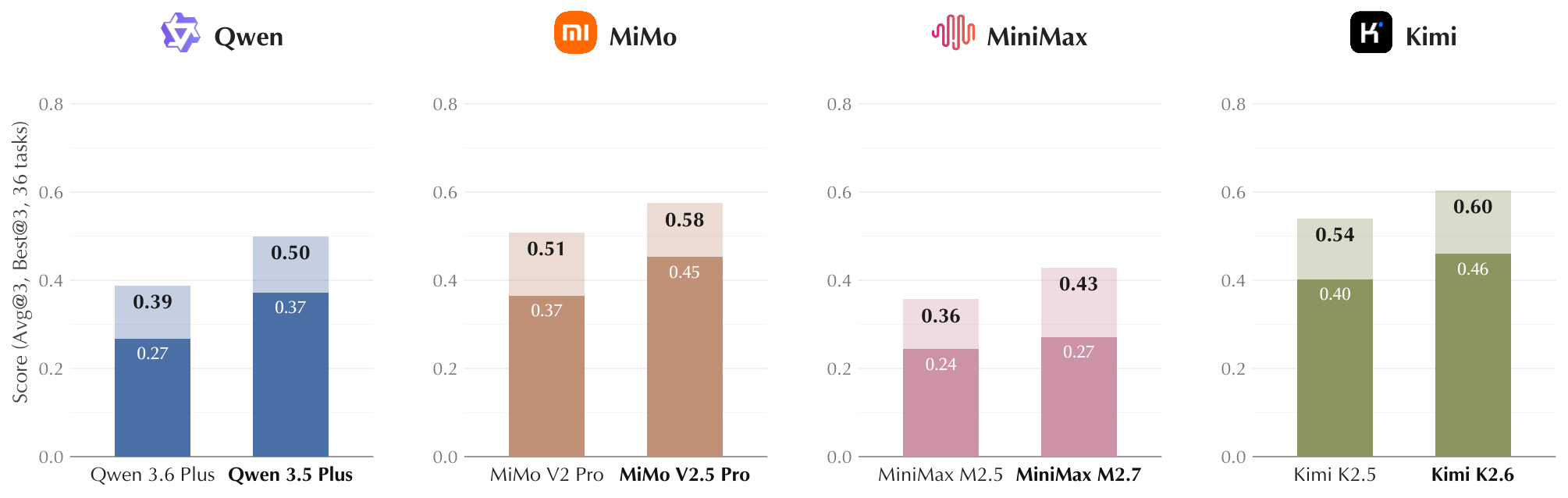}
\caption{Per-provider generation deltas on \bench{} across all 36 tasks. Three of the four pairs (MiMo, MiniMax, Kimi) exhibit modest gains from the older variant (left) to the newer one (right). Qwen is shown with the newer 3.6 Plus on the left to highlight its regression relative to 3.5 Plus}
\label{fig:model_generations}
\end{figure}

We next examine within-provider generation improvements while holding the harness at terminus-2. Figure~\ref{fig:model_generations} compares four old-to-new pairs: Qwen 3.5 Plus to 3.6 Plus, MiMo v2 Pro to v2.5 Pro, MiniMax M2.5 to M2.7, and Kimi K2.5 to K2.6.

Three of the four pairs show modest gains. MiMo improves the most, followed by MiniMax and Kimi. \textbf{Qwen 3.6 Plus is the only generation that regresses}, dropping $0.09$ on Avg@3 and $0.12$ on Best@3. This decline is consistent with the category breakdown in Table~\ref{tab:main}: while Qwen 3.6 Plus retains a strong Model Development score (0.88), its performance on CUDA, Puzzle \& Challenge, and System Optimization collapses to or near zero. 
We also observe that newer flagship models do not improve every category uniformly. For instance, MiniMax M2.7 improves overall but still trails the median on CUDA, while MiMo v2.5 Pro gains on Model Development yet loses ground on CUDA. Thus, provider-level generation lifts do not guarantee uniform gains across sub-domains.

\subsection{Stability Analysis}
\label{app:stability}

Across-trial stability is a distinct and important dimension from raw capability. A model that achieves $0.85$ on one trial but only $0.20$ on the other two will have the same Avg@3 as a model that consistently scores $0.65$ across all three trials, yet their reliability differs substantially. Reporting only the mean therefore masks critical variance. Table~\ref{tab:stability} quantifies across-trial dispersion for the 11 main-set models.

For each (model, task) pair, we compute four complementary stability metrics from the three independent trials: the mean per-task standard deviation $\bar{\sigma}$, the mean per-task range $\bar{R}$, the coefficient of variation $\text{CV}=\bar{\sigma}/\text{Avg@3}$, and the normalized dispersion $\bar{\sigma}/\text{Best@3}$. All four metrics show consistent trends.

\begin{table}[htbp]
\centering
\setlength{\tabcolsep}{6pt}
\renewcommand{\arraystretch}{1.10}
\begin{tabular}{lcccccc}
\toprule
\textbf{Model} & \textbf{Avg@3}\,$\uparrow$ & \textbf{Best@3}\,$\uparrow$ & $\bar{\sigma}$\,$\downarrow$ & $\bar{R}$\,$\downarrow$ & \textbf{CV}\,$\downarrow$ & $\bar{\sigma}/$Best\,$\downarrow$ \\
\midrule
\raisebox{-0.30ex}{\includegraphics[height=2.4ex]{logos/claude-opus-4.6.png}}~\texttt{claude-opus-4.6} & \textbf{0.77} & \textbf{0.85} & \textbf{0.099} & \textbf{0.180} & \textbf{0.13} & \textbf{0.12} \\
\raisebox{-0.30ex}{\includegraphics[height=2.4ex]{logos/mimo-v2.5-pro.png}}~\texttt{mimo-v2.5-pro} & 0.54 & 0.68 & 0.152 & 0.278 & 0.28 & 0.22 \\
\raisebox{-0.30ex}{\includegraphics[height=2.4ex]{logos/gemini-3.1-pro.png}}~\texttt{gemini-3.1-pro} & 0.57 & 0.68 & \underline{0.114} & \underline{0.209} & \underline{0.20} & \underline{0.17} \\
\raisebox{-0.30ex}{\includegraphics[height=2.4ex]{logos/kimi-k2.6.png}}~\texttt{kimi-k2.6} & 0.52 & 0.67 & 0.189 & 0.339 & 0.37 & 0.28 \\
\raisebox{-0.30ex}{\includegraphics[height=2.4ex]{logos/glm-5.png}}~\texttt{glm-5} & 0.52 & 0.66 & 0.155 & 0.279 & 0.30 & 0.23 \\
\raisebox{-0.30ex}{\includegraphics[height=2.4ex]{logos/gpt-5.4.png}}~\texttt{gpt-5.4} & 0.47 & 0.63 & 0.154 & 0.281 & 0.33 & 0.24 \\
\raisebox{-0.30ex}{\includegraphics[height=2.4ex]{logos/deepseek-v4-pro.png}}~\texttt{deepseek-v4-pro} & 0.45 & 0.60 & 0.189 & 0.344 & 0.42 & 0.31 \\
\raisebox{-0.30ex}{\includegraphics[height=2.4ex]{logos/grok-4-20.png}}~\texttt{grok-4-20} & 0.43 & 0.55 & 0.127 & 0.226 & 0.29 & 0.23 \\
\raisebox{-0.30ex}{\includegraphics[height=2.4ex]{logos/hy3-preview.png}}~\texttt{hunyuan-3-preview} & 0.38 & 0.53 & 0.175 & 0.313 & 0.46 & 0.33 \\
\raisebox{-0.30ex}{\includegraphics[height=2.4ex]{logos/minimax-m2.7.png}}~\texttt{minimax-m2.7} & 0.35 & 0.50 & 0.152 & 0.276 & 0.44 & 0.30 \\
\raisebox{-0.30ex}{\includegraphics[height=2.4ex]{logos/qwen3.6-plus.png}}~\texttt{qwen-3.6-plus} & 0.36 & 0.49 & 0.158 & 0.289 & 0.43 & 0.32 \\
\bottomrule
\end{tabular}
\caption{Across-trial stability for the 11 frontier models. $\bar{\sigma}$ denotes the mean per-task standard deviation, $\bar{R}$ the mean per-task range, CV the coefficient of variation ($\bar{\sigma}/\text{Avg@3}$), and the final column normalizes $\bar{\sigma}$ by Best@3. Lower values indicate higher stability. Best entry per column is in \textbf{bold} and runner-up is \underline{underlined}. Rows are ordered by descending Best@3 to match Table~\ref{tab:main}.}
\label{tab:stability}
\end{table}

Three observations stand out:

\textbf{(i) Stability and capability are correlated but not identical.} \texttt{claude-opus-4.6} is both the highest-scoring and the most stable model ($\bar{\sigma}=0.099$), exhibiting less than half the dispersion of other top-5 models. At the lower end, three of the four weakest models (\texttt{hunyuan-3-preview}, \texttt{minimax-m2.7}, \texttt{qwen-3.6-plus}) also show high variance (CV $\geq 0.43$). The relationship is not strictly monotonic: \texttt{gemini-3.1-pro} and \texttt{grok-4-20} are notably more stable than peers with similar Avg@3, while \texttt{kimi-k2.6} is mid-tier in performance but among the noisiest models.

\textbf{(ii) Single-trial evaluation is unreliable for high-variance models.} For models with CV $\geq 0.40$ (\texttt{deepseek-v4-pro}, \texttt{hunyuan-3-preview}, \texttt{minimax-m2.7}, \texttt{qwen-3.6-plus}), the mean across-trial range reaches $0.28$--$0.34$ on a $[0,1]$ scale. A single rollout can therefore land anywhere within roughly one-third of the full score range, making single-shot rankings highly unreliable. We recommend using Avg@3 (or more trials) as the primary metric for such models.

\textbf{(iii) Best@3 over-credits noisy models.} The gap between Best@3 and Avg@3 widens with increasing $\bar{\sigma}$ (Pearson $r=0.84$). Relying solely on Best@3 therefore inflates the apparent capability of high-variance models more than that of stable ones, which is why we report both metrics in the main leaderboard.

\end{document}